\documentclass{article}

\usepackage{arxiv}
\usepackage{wrapfig}

\usepackage[utf8]{inputenc} 
\usepackage[T1]{fontenc}    
\usepackage{hyperref}       
\usepackage{url}            
\usepackage{booktabs}       
\usepackage{amsfonts}       
\usepackage{nicefrac}       
\usepackage{microtype}      
\usepackage{lipsum}
\usepackage{graphicx}
\graphicspath{ {./images/} }

\usepackage[font=small,labelfont=bf]{caption}

\title{End-to-end deep learning for directly estimating grape yield from ground-based imagery}

\author{
 Alexander G. Olenskyj \\
  Dept. of Bio. and Ag. Engineering\\
  AI Inst. for Food Systems (AIFS) \\
  Univ. of California, Davis\\
   \And
 Brent S. Sams \\
  Dept. of Winegrowing Research\\
  E\&J Gallo Winery\\
  \And
 Zhenghao Fei \\
  Dept. of Bio. and Ag. Eng.\\
  AI Inst. for Food Systems (AIFS) \\
  Univ. of California, Davis\\
  \And
 Vishal Singh \\
  Dept. of Mech. and Aero. Eng.\\
  Univ. of California, Davis\\
  \And
 Pranav V. Raja \\
  Dept. of Bio. and Ag. Eng.\\
  AI Inst. for Food Systems (AIFS) \\
  Univ. of California, Davis\\
  \And
 Gail M. Bornhorst \\
  Dept. of Bio. and Ag. Eng.\\
  Univ. of California, Davis\\
  Riddet Inst., Palmerston North, NZ \\
  \And
 J. Mason Earles \\
  Dept. of Viticulture and Enology\\
  Dept. of Bio. and Ag. Eng.\\
  AI Inst. for Food Systems (AIFS) \\
  Univ. of California, Davis\\
  \\
  }

\begin{document}
\maketitle
\begin{abstract}
Yield estimation prior to harvest is a powerful tool in vineyard management, as it allows growers to fine-tune management practices to optimize yield and quality. However, yield estimation is currently performed using manual sampling, which is time-consuming and imprecise. This study demonstrates the applicability of nondestructive proximal imaging combined with deep learning for yield estimation in vineyards. Continuous image data collection using a vehicle-mounted sensing kit combined with collection of ground truth yield data at harvest using a commercial yield monitor allowed for the generation of a large dataset of 23,581 yield points and 107,933 images. Moreover, this study was conducted in a commercial vineyard which was mechanically managed, representing a challenging environment for image analysis but a common set of conditions in the California Central Valley. Three model architectures were tested: object detection, CNN regression, and transformer models. The object detection model was trained on hand-labeled images to localize grape bunches, and detections were either counted or their pixel count was summed to obtain a metric which was correlated to grape yield. Conversely, regression models were trained end-to-end to directly predict grape yield from image data without the need for hand labeling. Results demonstrated that both a transformer model as well as the object detection model with pixel area processing performed comparably, with a mean absolute percent error of 18\% and 18.5\%, respectively on a representative holdout dataset. Saliency mapping was used to demonstrate the attention of the CNN regression model was localized near the predicted location of grape bunches, as well as on the top of the grapevine canopy. Overall, the study demonstrated the applicability of proximal imaging and deep learning for prediction of grapevine yield on a large scale. Additionally, the end-to-end modeling approach was able to perform comparably to the object detection approach while eliminating the need for hand-labeling.
\end{abstract}


\section{Introduction}
Yield estimation is a valuable tool for crop management and precision agriculture. Estimation of crop yield in advance of harvest allows growers to make more informed decisions in negotiating pricing contracts or allocating quantities of crop for sale, and may also inform crop management decisions \cite{Diago2012,Nuske2014}. There are many approaches which can be taken towards estimating crop yield, the most common method involving manually sampling from a very small percentage (e.g., approximately 1\%) of plants and extrapolating the yield data to the entire field. While this process is comparatively low-cost, it can be both time-consuming and imprecise \cite{Liu2020}. As opposed to manual sampling, remote sensing can capture information from an entire field in a short period of time, whether via satellite or via unmanned aerial vehicles (UAVs). The advantages of satellite-based remote sensing are speed, coverage, and a lack of requirement for manual labor. However, the resolution of publicly available satellite-based remote sensing is poor, typically between 10 and 30 m2 \cite{Khaliq2019}. Atmospheric conditions can also be an issue as clouds, smoke, and haze can obscure the area of interest. Additionally, data collection cannot be scheduled, due to the regular orbit of the satellite. Remote sensing images acquired using UAVs can provide increased resolution, but requires a trained pilot \cite{Khaliq2019,Yang2019}. Additionally, for specialty crops where fruit is located beneath vegetative cover, such as tomatoes and grapes, the overhead angle of images creates difficulties in imaging, especially when foliage is dense \cite{DiGennaro2019}.
Proximal imaging, referred to here as imaging from ground-based equipment, has also been extensively explored for yield estimation \cite{Gongal2015}. Proximal sensing using imagery can be advantageous, as high-resolution images from a lateral perspective can be collected from underneath a canopy. However, like UAV imaging, proximal imaging requires specialized equipment. Additionally, while the resolution and perspective of proximal imagery can allow for direct visualization and quantification of fruit, occluded fruits can lead to inaccuracy in the estimation \cite{Dunn2004,Mu2020,Wang2021}. Nevertheless, there has been extensive work in image-based proximal sensing for fruit detection and counting \cite{Gongal2015}. 
Recently, modeling techniques incorporating vision-based machine learning, which have demonstrated success in numerous fields, have seen considerable use in fruit detection and counting via proximal imaging \cite{Bargoti2017,Gene-Mola2019,Santos2020}. Much of the previous work has been performed in orchard crops, such as apples \cite{Bargoti2017,Hani2020}, oranges \cite{Maldonado2016}, and mangos \cite{Payne2014}. Studies in other specialty crops, including tomatoes \cite{Mu2020,Rahnemoonfar2017a} and grapes \cite{Li2021,Liu2017b,Liu2015b,Santos2020,Sozzi2022} have also been conducted. In vineyards, much of the existing literature in yield estimation from proximal imagery have consisted of methods leveraging feature engineering and computer vision to count individual grape berries \cite{Millan2018,Nuske2014,Rose2016}, although pixel count has also been related to grapevine yield \cite{Diago2012}.
In very recent years, studies on proximal imaging for fruit detection have moved away from hand-crafted feature and algorithm development towards deep learning methods \cite{Gene-Mola2019,Li2021,Milella2019,Santos2020,Sozzi2022}. Deep learning methods used for fruit detection reduce the bias involved in model development via their flexibility in learning both the optimal features and optimal relationships between features to converge to the desired result. However, while these methods are robust to a small amount of occlusion, they cannot account for completely occluded fruits \cite{Gongal2015,Mu2020}. In an effort to overcome the issue of occlusion, previous research in grape yield estimation has demonstrated some benefit of incorporating variable grape visibility into yield estimation models for improving performance using statistical modeling \cite{Millan2018,Nuske2014}. However, these approaches have involved tuning the model to the specific dataset, either by incorporating previous years’ data or specifying expected properties of the image data, such as mean berry area.
While previous yield estimation methods have focused primarily on semantic segmentation, object detection, and object masking used for fruit detection, deep learning methods have also been used to map image data directly to more abstract outputs, such as image captions, monocular depth estimates \cite{Hu2019}, scalar values of food calories \cite{Ege2017}, subject age \cite{Othmani2020}, and more \cite{ZakirHossain2019}. Once training is completed, these deep learning models map relationships from semantically complex combinations of image features to accurate predictions in unrelated modalities. This “end-to-end learning” approach can be applied to yield estimation by directly relating images and desired yield information, as opposed to the use of vegetative indices or models for detection or masking. This hypothesis was tested with promising results in the field of UAV-based yield estimation, in which a convolutional neural network (CNN) trained to forecast rice grain yield directly from high-resolution RGB images outperformed an NDVI-based approach \cite{Yang2019}. End-to-end learning has also been demonstrated in the context of grape yield estimation from proximal imagery, where a regression CNN model was used to directly predict yield in units of mass from images \cite{Silver2019}. However, images from that previous study were collected individually with a smartphone, which did not represent a scalable method. Additionally, vines were prepared specifically for imaging via placement of a calibration marker in each frame. Finally, extensive manual processing and cropping of the images was performed. Nevertheless, the study represents a proof-of-concept for utilization of DL in viticultural yield estimation. Like the improvements garnered by deep learning-based methods in learning features, as opposed to relying on prior assumptions, end-to-end methods used to directly relate images and yield may also be able to learn features from images other than solely those which contribute to fruit localization. For example, information such as canopy structure, fruit position, or other features may be relevant for the purpose of calibrating the yield estimate to account for occluded fruit. In addition to the above advantages garnered by end-to-end modeling, end-to-end models do not require labeling of objects in image data. Hand-labeling is a universal bottleneck in object detection studies and eliminating the labeling step can allow for the use of more data.
In this study, three different deep learning methods were compared in their ability to estimate yield in vineyards at varying levels of spatial resolution. Yield values in metric tons per hectare (t/ha) acquired during harvest were predicted using a fruit detection model as well as two end-to-end networks trained to predict yield directly from images: a CNN and a transformer network. The transformer architecture has recently been shown to perform well on a wide range of tasks, including natural language processing, as well as many vision-based tasks such as classification, segmentation, and object detection \cite{Carion2020,Dosovitskiy2020,Vaswani2017,Xie2021}. Considering the design of the transformer architecture in accepting sequences of data, such as words or image patches, this flexible architecture was applied to this dataset to allow for a set of neighboring images recorded near a ground truth yield measurement to be accounted for in the estimation.
In addition to demonstrating the application of novel modeling techniques, this study also incorporated a uniquely large dataset. Previous datasets used for model development have consisted of as few as 10 vines \cite{Diago2012} to up to 1,212 vines \cite{Nuske2014}. Yield data collected by hand-weighing grapes is typically used as a ground truth measure. However, yield monitors which generate high-resolution yield data as crops are harvested are commonly used by growers to map yield on a large scale. These monitors have been used previously in conjunction with remote sensing studies for yield estimation in crops such as sorghum and cotton \cite{Yang2004} as well as grapes \cite{Sun2017}. However, proximal imaging studies on larger datasets of grape yield are still lacking, primarily due to the expensive nature of data collection. In this study, models were trained and evaluated on a dataset containing 23,581 yield values measured at harvest using a yield monitor. To the best of the authors’ knowledge, this study represents the first application of end-to-end modeling using minimally processed images collected from a moving platform for grape yield estimation. Additionally, this study represents the one of the first applications of vision transformers in direct yield estimation from image data. Furthermore, the scale of the yield data considered in the present work was an order of magnitude greater than in previous studies. Finally, the data in this work were collected from a commercial, mechanically managed vineyard, as opposed to previous studies conducted in vineyards with manual management. Mechanized vineyards generally consist of increased occlusion, whereas manual management allows for better targeting of specific shoots and leaves as compared with machine implements.

\section{Materials and Methods}
\label{sec:MatMethods}

\subsection{Field Layout}

\begin{table}
  \small
  \centering
  \caption{Vine characteristics and data acquired in each block.}
    \begin{tabular}{rrrrrrrlrr}
    \toprule
    \multicolumn{1}{p{2.215em}}{Block} & \multicolumn{1}{p{4.215em}}{Training method} & \multicolumn{1}{p{4em}}{Trellis system} & \multicolumn{1}{p{3.215em}}{Rootstock} & \multicolumn{1}{p{3.215em}}{Vine Spacing (m)} & \multicolumn{1}{p{3.215em}}{Row Spacing (m)} & \multicolumn{1}{p{4.215em}}{Approx. Vines Measured} & \multicolumn{1}{p{4.215em}}{Dataset} & \multicolumn{1}{p{3.215em}}{Yield Points} & \multicolumn{1}{p{3.215em}}{Images} \\
    \midrule
          &       &       &       &       &       &       & Training & 5343  & 35279 \\
    1     & \multicolumn{1}{l}{Stacked-T} & \multicolumn{1}{l}{Sprawl} & \multicolumn{1}{l}{So4} & 1.5   & 3.4   & 7271  & Validation & 987   & 6868 \\
          &       &       &       &       &       &       & Test  & 1264  & 9168 \\
          &       &       &       &       &       &       & Training & 1241  & 11398 \\
    2     & \multicolumn{1}{l}{Quadrilateral} & \multicolumn{1}{l}{Sprawl} & \multicolumn{1}{l}{So4} & 1.2   & 3.4   & 2201  & Validation & 208   & 2478 \\
          &       &       &       &       &       &       & Test  & 357   & 3625 \\
          &       &       &       &       &       &       & Training & 315   & 3540 \\
    3     & \multicolumn{1}{l}{High Wire} & \multicolumn{1}{l}{Sprawl} & \multicolumn{1}{l}{1103P} & 1.8   & 3     & 1173  & Validation & 54    & 793 \\
          &       &       &       &       &       &       & Test  & 119   & 1125 \\
          &       &       &       &       &       &       & Training & 2610  & 19656 \\
    4     & \multicolumn{1}{l}{Quadrilateral} & \multicolumn{1}{l}{Sprawl} & \multicolumn{1}{l}{So4} & 1.5   & 3.4   & 2535  & Validation & 606   & 4580 \\
          &       &       &       &       &       &       & Test  & 847   & 6565 \\
    \bottomrule
    \end{tabular}%
  \label{tab:table1}%
\end{table}%

\begin{figure}
  \centering
  \includegraphics{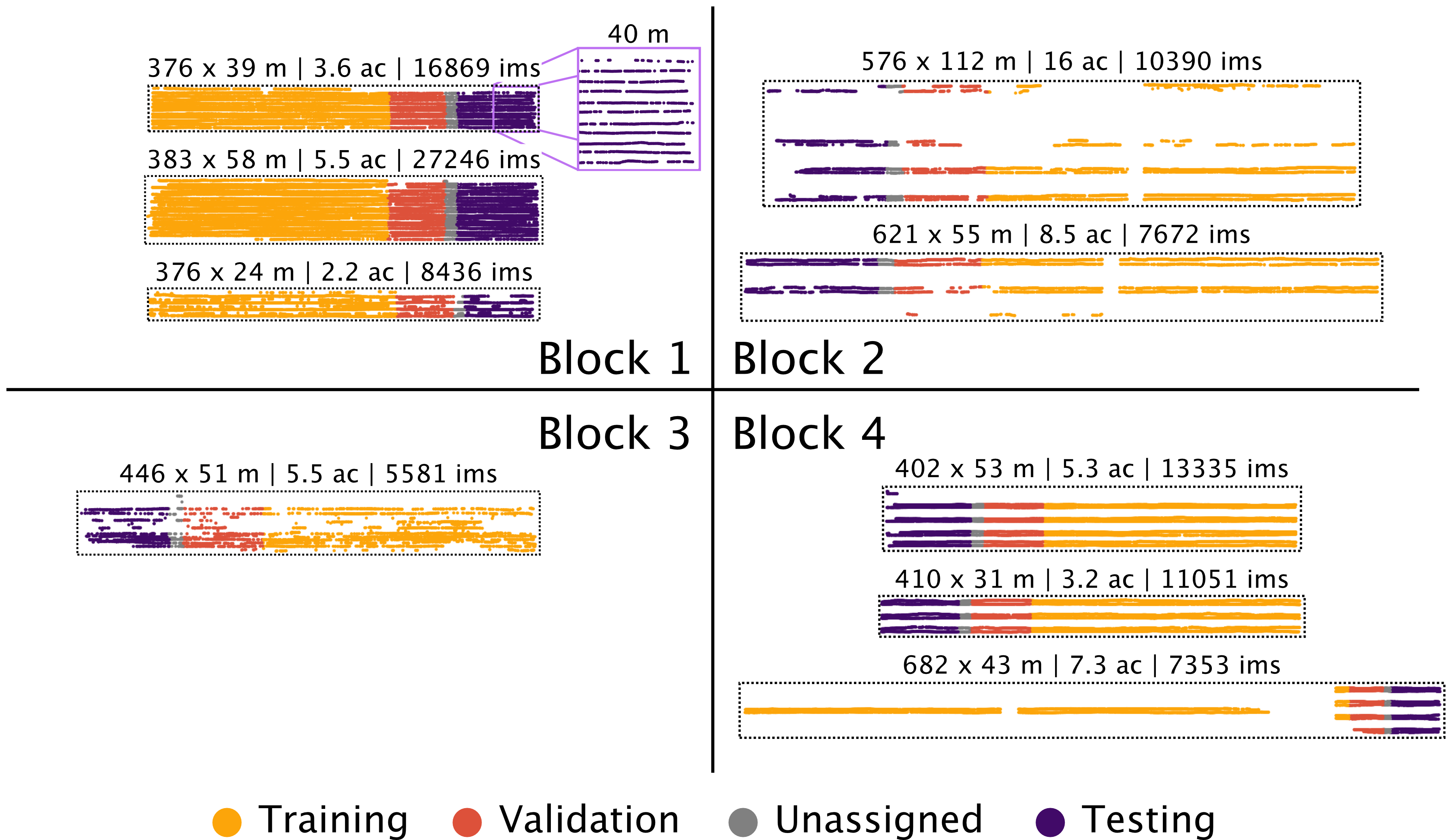}
  \caption{Image locations used in this study, colored according to their split in training, validation, and testing sets. Points located between validation and testing sets were labeled as unassigned and not used in the training process to minimize similarity between training and validation set images to the test set. Dimensions, acreage, and number of images are given above each boxed region.}
  \label{fig:fig1}
\end{figure}

Data were collected in a commercial grape vineyard (Vitis vinifera L. cv. Cabernet Sauvignon) within the California Central Valley region in September 2020. Image data were collected approximately three weeks before harvest, while ground truth data was collected during harvest. The vineyard was divided into four blocks representing differences in management and/or trellis type (Fig. \ref{fig:fig1}). For the purposes of this study, block was not considered during the training of any model. However, data are presented as separated by block to demonstrate the effect of management and vine characteristics on model performance. Table \ref{tab:table1} contains details about vines within each block. Grapevine rows were arranged in an East-West configuration for all blocks.

\subsection{Data Acquisition}
\subsubsection{Image Data}

\begin{wrapfigure}{L}{0.55\textwidth}
  \centering
  \includegraphics{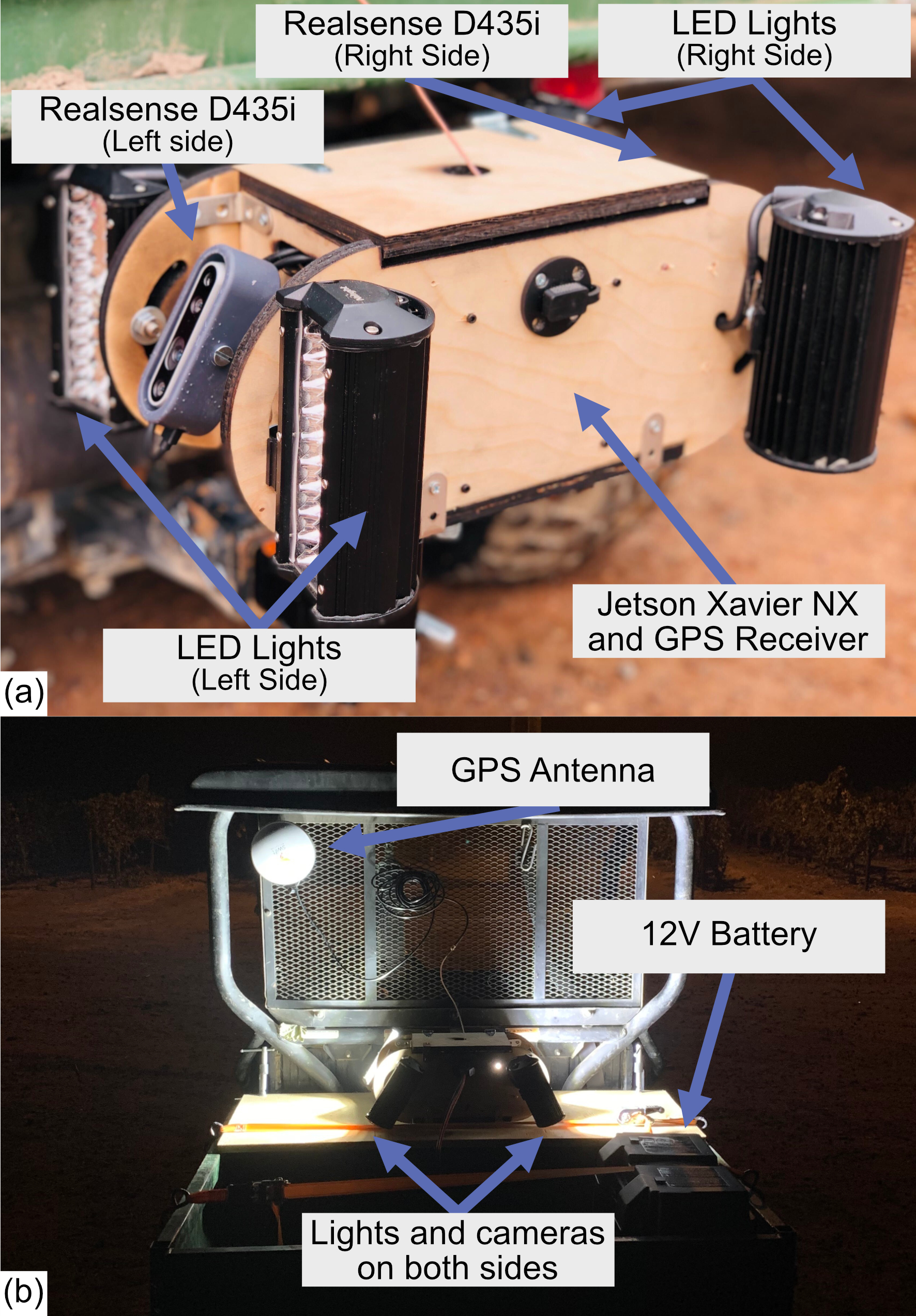}
  \caption{Sensing kit designed at UC Davis used for image data collection. a) detailed view of the sensing kit. b) depiction of the position of the sensing kit during imaging for this study.}
  \label{fig:fig2}
\end{wrapfigure}

Image data acquisition was performed using a low-cost sensing kit (Fig. \ref{fig:fig2}) with two RealSense D435i cameras (Intel, Santa Clara, CA) vertically oriented opposite each other, perpendicular to the direction of travel. Location data was collected using a Piksi Multi GPS module and antenna (Swift Navigation, San Francisco, CA). In addition to the sensing modules, two 120W LED arrays on each side of the kit were included to illuminate the environment (Nilight, Englewood, NJ). All components were managed by a Jetson Xavier NX single board computer (NVIDIA, Santa Clara, CA) using the ROS platform. This sensing kit was attached to the back of an agricultural utility vehicle and powered by a deep cycle battery during data acquisition. Imaging was performed at night to control illumination consistency. RGB and depth image data were collected simultaneously from both cameras at 15 Hz frequency, although only the RGB imagery was used for this study as depth images from the RealSense camera rely on infrared sensors, which provided poor signal at night. The cameras implement an infrared projector to assist with depth estimation, but due to the heterogeneous texture of the grapevine foliage in the images, the reconstructed depth maps were of low quality and were not used for further analysis. Images were compressed using JPEG compression to reduce storage requirements. Satellite-based augmentation system corrected GPS data was collected at 10 Hz. Although the GPS module on the sensing kit is capable of real-time kinematic (RTK) correction, a base station was not available during imaging and therefore only satellite-based corrections were used. Image georeferencing was performed based on timestamp matching. The average speed of the sensor during imaging was 3.3 m/s, providing a 0.22 m spatial resolution for image data before image filtering (Section 2.3.2).

\subsubsection{Ground Truth Data}
Grapes were mechanically harvested using a commercial harvester fit with a load cell along with a GPS unit for yield monitoring (Advanced Technology Viticulture, Adelaide, Australia). Force data from the load cell were recorded at with a spatial resolution of approximately 0.77 meters along each row. Data were collected continuously and the yield monitor was calibrated with a scalar time offset such that the location recorded for each yield value best represented the location where the grapes were grown.

\subsection{Data Handling}
\subsubsection{Ground Truth Data}
Yield monitor data was first filtered manually to remove artifacts, such as yield points from gaps in rows (where recorded yield appeared artificially lowered), spurious GPS points from outside of the rows, and rows in which the frequency of yield points was erroneously low. This was performed by plotting yield points based on their geospatial location and coloring by yield. Points meeting the above criteria were considered artifacts resulting from noise present in either the GPS system or the load cell and were manually removed. This processing was done prior to image data association or model development. In total, 10.5\% of points were removed. Data were then filtered further to remove outliers, defined as points with yield of more than 1.5 interquartile range values away from the first and third quartiles \cite{Tukey1977}. Next, data from the yield monitor was calibrated based on the total yield from each vineyard as weighed by the receiving winery. Yield density was calculated by dividing the mass of grapes per block (tons) by the area of each block (hectares). All yield points within each block were then scaled proportionally such that the mean value was equivalent to the mean yield density measured in each block using a commercial scale as part of routine operations. This methodology was similar to that performed by Sun et al. \cite{Sun2017}, and similar to the methodology used in the grape industry.

\subsubsection{Image Data}
The total ground-imagery dataset before filtering and data association consisted of 274,944 images. Image data were filtered to remove poor quality images, which consisted of images where a shoot extending into the inter-row space obscured the camera, images recorded during turns between rows, and excessively blurred images where individual grape clusters could not be seen clearly. This was done by randomly selecting 3,000 images from the dataset and manually labeling them with binary labels (“keep” and “toss”). This subset of labeled data was split into training, validation, and test sets with 1800, 600, and 600 images, respectively, and a MobileNetV2 model was trained to filter images \cite{Sandler2018}. The trained model achieved an accuracy of 91.2\% on the test set and was used to filter the rest of the dataset.

\subsubsection{Association of Image and Yield Monitor Data}
\begin{figure}
  \centering
  \includegraphics[width=1.0\textwidth]{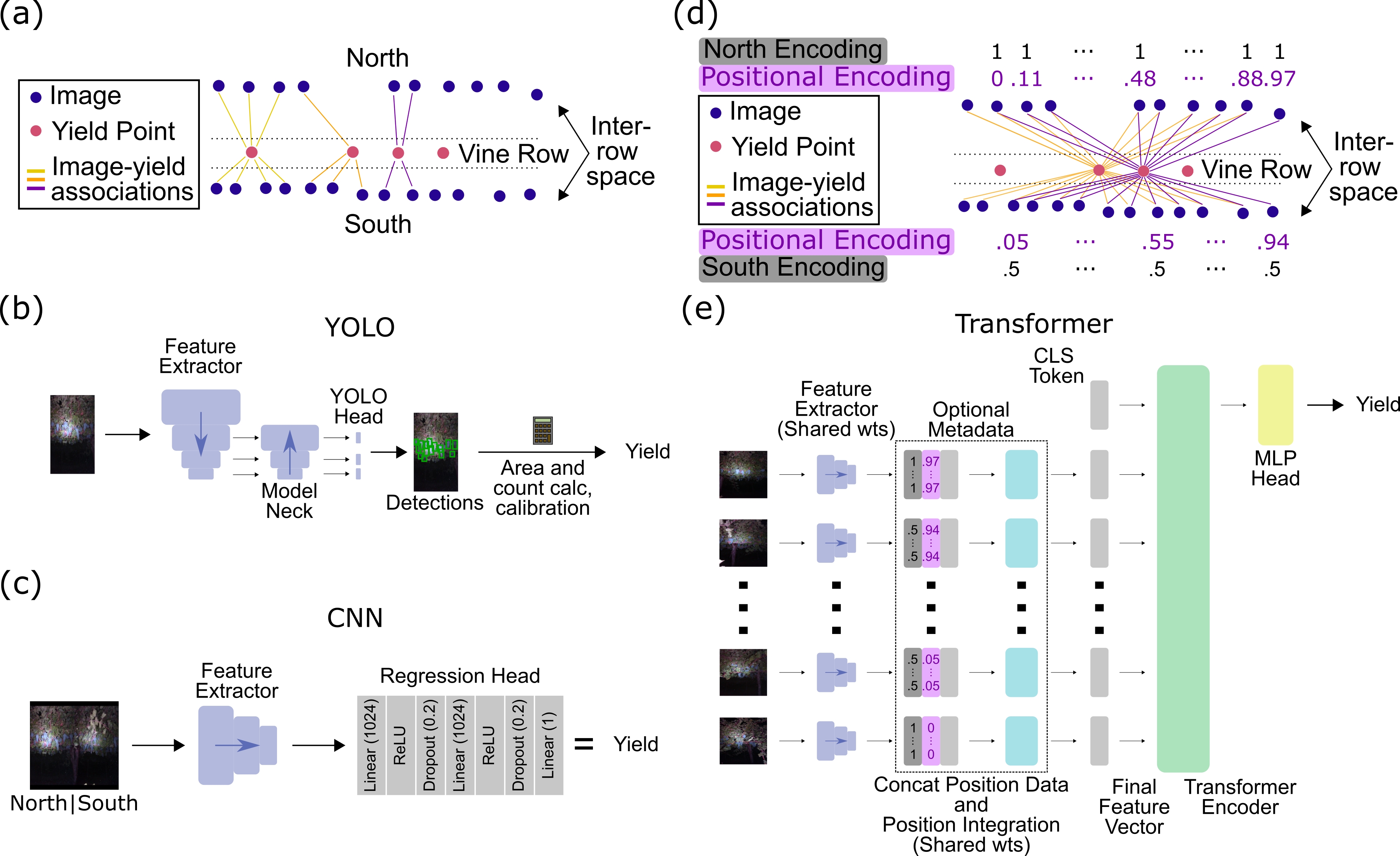}
  \caption{Data handling and model architectures. a) Images in object detection and CNN models were associated with their closest yield point (three yield points shown with image-yield associations in different colors). b) The YOLOv5 model was applied to individual frames and detections were processed to calculate yield values at each point. c) The CNN architecture was applied to a pair of images, one from each side of the vine. The final scalar output represented the predicted yield. d) The transformer model was associated a window of image points with each yield point (two yield points shown with image-yield associations in different colors). Positional encodings were extracted based on relative location of image and yield points. e) The transformer architecture. Each image was passed through the same feature extractor. When used, positional encoding was performed after feature extraction.}
  \label{fig:fig3}
\end{figure}

Yield data was collected during harvest, and each yield point was marked directly over the grapevine row, with a different spatial resolution as compared with image data. Therefore, association of yield and image data was necessary (Fig. \ref{fig:fig3}). For association of each image with a corresponding yield measurement, a distance matching algorithm was designed to match each image with its closest yield point, considering the orientation of the camera (North or South) for each image (Fig. \ref{fig:fig3}a). For training the transformer model (Section 2.4.3), all measured yield points with at least one North-facing and at least one South-facing image within 5 m to the east or west of the measured yield point were kept (Fig. \ref{fig:fig3}d). Any yield point without images from both the North and South side of the vine were discarded. This set of measured yield points was further pruned for training the CNN model (Section 2.4.2), where yield points were only retained if they were associated with at least one image from each side of the vine that was not closer to an adjacent yield point (Fig. \ref{fig:fig3}a).

In total, this left 23,581 yield points in the dataset used to train the transformer model, and 14,302 yield points in the dataset used to train the CNN model. Yield points used to train the transformer model represented approximately 13,180 vines. This value was estimated by rounding all collected yield points to the nearest interval representing vine spacing along the direction of the rows and row spacing along the perpendicular direction within each block. Yield points with the same rounded values were considered to come from the same vine. After removing poor quality images, images far away from yield points, and images associated with yield points for which both sides of the vine were not accounted for, these yield points were associated with 107,933 and 80,009 images in the transformer and CNN model sets, respectively.

\subsubsection{Dataset splitting for model development}
Yield values were split into training, validation, and testing sets according to their location, such that no training or validation data was adjacent to a point in the test set. This was performed to reduce the influence of spatial autocorrelation on model performance (Fig. \ref{fig:fig1}). 
Within the dataset used to train the transformer model, this resulted in 15,024 yield points in the training set, 3,354 in the validation set, and 4,529 in the test set. Yield points and associated image counts by dataset and block can be found in Table \ref{tab:table1}. For the CNN and object detection models, this resulted in 9,509 yield points in the training set, 1,855 in the validation set, and 2,587 in the testing set.

\subsection{Model Architectures and Training Procedures}
\subsubsection{Object Detection}

The object detection model required labeled bounding boxes around grape clusters visible within a subset of images. Specifically, 150 images were selected at random from the training and validation sets of the dataset. Grape bunches in these images were labeled with bounding box labels. Care was taken to only label bunches on the near side of the vine, in the event that bunches from the far side of the canopy were visible in the image. For vines trained with quadrilateral trellises, grapes on the near side were defined as grapes on the closer set of cordons. For vines trained with bilateral trellises, the cordons were used as a dividing plane, where grape bunches on the near side of the cordons were selected. These 150 images were divided into training, validation, and testing sets of 98, 24, and 28 images respectively, considering the distribution of images from each management block.
The object detection model was trained with YOLOv5 \cite{Jocher2021}. Images were resized to 640 x 640 pixels and augmentation was performed using the mosaic method along with variation of hue, saturation, and luminance values representing the default augmentations. The model was trained for 300 epochs. The best model was determined using a weighted average of a) 0.9 times mean average precision (mAP) averaged over intersection over union (IOU) values of 0.5:0.95 in increments of 0.05 and b) 0.1 times mAP at an IOU of 0.5 only. This evaluation metric was computed on the validation set after each epoch.
Bounding box count (i.e. bunch count) and bounding box area per image were used in yield estimation. For each location with yield data, mean bunch count or mean summed bounding box area was determined for all images on each side of the vine. The two sides were then summed to obtain the final estimate of bunch count or area (px\textsuperscript{2}). Bunch count and area were then correlated with the yield values in all training set images from the full dataset. A simple linear regression with the intercept fixed at the origin was used to relate bunch count and area with yield. Evaluation on the test set was conducted by first obtaining mean count or summed area over both sides of the vine, then converting the value to yield with the corresponding linear fit.

\subsubsection{Convolutional Neural Network Model Architecture}
For the CNN model, the ResNet18 architecture \cite{He2016} was adapted to a regression approach (Fig. \ref{fig:fig3}c). The 1000-length final linear layer of the original architecture was replaced with a regression head composed of two 1024-length linear layers, each followed by a ReLU and 20\% dropout step \cite{Nair2010,Srivastava2014}. A final linear layer of size 1 was used to represent the yield corresponding with the input. No softmax or other post-processing was performed. Weights for the model were randomly initialized. Input data consisted of one frame from each side of the vine concatenated horizontally (Fig \ref{fig:fig3}c). Since multiple frames on either side of the row were associated with a given yield point, the North side and South side image used in the input were randomly selected from the available images associated with each point during training. The validation set consisted of seeded random selection of frames, such that the North and South frames were consistent between epochs. 
The model was trained using an adaptive loss function \cite{Barron2019} for 1.97M steps of batch size 12, which were divided into 25 epochs. North and South frames were augmented separately using random horizontal flips, and random median blurring. The North image was always on the left side of the input. For inference on the test set, all combinations of North and South image pairs were input to the model for each yield point, and the average predicted yield was taken. Model weights were saved and a validation score was computed after each epoch. The model with the lowest validation loss was selected for performance evaluation.

\subsubsection{Transformer Model Architecture}
While concatenation of two images was used to increase the context of the data with respect to the location at which yield data were measured by providing two views of the vine in the CNN approach, concatenating additional frames would quickly exceed available GPU memory or require substantial image downscaling leading to information loss. The vision transformer architecture employed in this study accounts for this limitation by accepting a sequence of images as its input. Additionally, due to the attention mechanisms in the model architecture, the influence of each input image to the final predicted value is allowed to vary \cite{Dosovitskiy2020,Vaswani2017}. This was of particular importance in this work, due to the use of a mechanical yield monitor in lieu of hand counting or weighing fruit. While the yield monitor allowed for generation of a uniquely large dataset, both the magnitude as well as the positional accuracy of the yield measurements were compromised by the continuous nature of data collection. Specifically, harvested grapes were measured with a load cell after a small delay, during which the grapes were conveyed from the vine to the instrument. While this time was accounted for with a scalar offset in this study, other factors such as the mass flow rate of fruit in and out of the harvester machinery have been shown to affect this value slightly \cite{Searcy1989}. As such, the yield values reported in this study were more appropriately derived from a distribution of vines surrounding the marked position. Therefore, a transformer model with the capacity to accept data from areas surrounding the location of the measured yield point was particularly well-suited to the present task. Specifically, predictions at each yield point were made from all images recorded within a window of 5 m to the east and west of the yield point (10 m total window size). Unlike in the CNN approach, where each image was only associated with a single yield point, the sliding window approach used for the transformer model allowed for the possibility that an image may be associated with multiple yield points if the image was located within 5 m of more than one yield point (Fig. \ref{fig:fig3}d). In these cases, these images were used multiple times as inputs to predict yield from each yield point (Fig. \ref{fig:fig3}d-e). A median of 47 images were associated with each yield point. This method of data handling was done to align with the method of ground truth data collection, where mixing may have occurred in the harvester as yield was being recorded, potentially allowing adjacent vines to influence the recorded yield in a given location.
The transformer model was based on the Vision Transformer (ViT) architecture \cite{Dosovitskiy2020} with modifications (Fig. \ref{fig:fig3}e). Instead of linearized image patches, token vectors input to the encoder were represented by ResNet34 features. Each image passed through the ResNet model, and the final set of activation maps was used for feature extraction. During training, ResNet weights were shared across all input images. Maps were averaged along the filter dimension, then linearized to generate a 256-length feature vector for each image.
In addition to the token representation, the positional encoding was also modified from its original format. Typically, positional encoding consists of sinusoidal or learned embeddings \cite{Vaswani2017}, which have been shown to be effective for inputs derived from uniformly cut patches of an image or language inputs. However, for both patch and word-based input, the spacing between inputs is consistent and only the relative location between inputs is important. In this application, the relevant positional information consisted of the distance between the measured yield point and the image location, as well as the orientation of the camera with respect to the vine. As a result, the prior information regarding the distance between each frame and the yield point, as well as the side of the vine represented by each image was used as the positional embedding. Setting the location of yield point at 0.5, locations 5 m to the east and west were scaled to 0 and 1, respectively (Fig. \ref{fig:fig3}d). This scalar was assigned to each input image and represented its position relative to the center frame at 0.5. Likewise, each input image was assigned either 0.5 to represent an image from the South side of the vine, or 1 to represent the north side. The scalar values encoding position and direction were repeated such that a vector the same size as the feature vector from the ResNet extractor was produced. Finally, as opposed to the typical summation of a positional vector with the token vector, a 1D convolutional layer was added to allow for a linear combination of each value in the 256-length feature vector with each of the scalar positional values to be learned and used to improve performance. Weights for the 1D convolution were shared across all input images. The inclusion of this positional metadata was performed to allow the model to selectively attend to images based on both their content as well as their position. To determine the influence that this added positional information had in an agricultural system, the model was trained both with and without information regarding position and orientation of each image frame (this optional step is represented in a dashed box in Fig. \ref{fig:fig3}e).
The remainder of the model architecture was minimally changed from the ViT architecture. The encoder was designed with a depth of 2 encoding blocks and 8 attention heads. The class token representing the entire input sequence was used as the input to the decoder to generate a prediction after the transformer encoder. The multilayer perceptron decoder used for classification was modified to output a single value to serve as the scalar prediction of yield. The model was trained with a mean squared error objective function for 50,000 steps and a batch size of 6, with gradient accumulation across every two batches for an effective batch size of 12. Training was divided into 20 epochs. Model weights were saved and validation performance was measured after each epoch. The weights with the lowest validation error were kept for performance evaluation.

\subsection{Performance Evaluation}
Object detection performance within the labeled dataset was measured using area under the precision-recall curve (AP) at an intersection over union of 0.5, as well as R\textsuperscript{2} and root mean squared error (RMSE) between labeled and predicted bunch counts and bounding box area.
Performance of each model on the test set was evaluated using RMSE, mean absolute percent error (MAPE), and R\textsuperscript{2} metrics between predicted and measured data. In order to compare all models, only the yield points represented in the test set of the CNN model were used for evaluation (Section 2.3.3). Therefore, all models were evaluated on a test set of 2,737 yield points. Additionally, due to the tendency for some of the trained models to predict values closer to the mean, as opposed to higher and lower yield values seen less frequently, the range expressed by each approach was also calculated using Eqn. 1

\begin{equation}
\frac{max(predicted\ yield) - min(predicted\ yield)}{max(measured\ yield) - min(measured\ yield)}
\end{equation}

Finally, to compare performance with remote sensing-based studies, post-hoc analysis was performed after model inference by spatially aggregating yield points and associated ground truth and predicted data into 10- or 20-meter square bins within each block. This was done by dividing the test dataset into separate zones that were densely populated with yield points, then binning points in the zones into grids of either 10- or 20-meter length, beginning at the lower left point of each zone. Within the test set, 179 bins were used at 10 m spacing, and 47 bins were used at 20 m spacing.

\subsection{Saliency Mapping and Model Visualization}
To gain insight into the features of the image used by the model to predict yield, the technique of Gradient-Based Class-Activation Mapping (Grad-CAM) was leveraged for the CNN model \cite{Selvaraju2016}. Typically used for classification tasks, this technique allows for visualization of the regions of the input image which contribute most strongly to increasing the value of an output class. As the model only outputs one value, the predicted yield, the visualization can be interpreted as the regions of the input image which contribute to raising the predicted yield value. While these regions are hypothesized to be localized in regions of input images with visible grape bunches, other features of the input, such as canopy density or shoot position, may also be relevant. 
The Grad-CAM approach was used on all images in the train set as well as the test set (all combinations of images from each side of the vine) to produce heatmaps scaled from 0 to 1. These maps were then averaged within each yield point, then averaged again over the entire dataset, giving equal weight to all yield points regardless of the number of associated images. For comparison with the Grad-CAM visualization, a heatmap was generated in a similar fashion for the train and test sets of the object detection model, where detected clusters in each image were assigned a value of 1 on a blank canvas of a size equal to the image size. These images, like the Grad-CAM heatmaps, were averaged within each yield point separately for the north and south sides of the vine, then averaged over all yield points to generate a similar heatmap to the CNN Grad-CAM plot.
Finally, in addition to visualization of model outputs, the scale of the dataset in this work was sufficient to produce yield maps. Maps were generated by aggregating yield points spatially at 10 m resolution, then plotting the aggregated regions on a map colored by yield value.

\section{Results and Discussion}
\subsection{Object Detection Approach}
\subsubsection{Internal Validation}

\begin{wrapfigure}{L}{0.5\textwidth}
  \centering
  \includegraphics{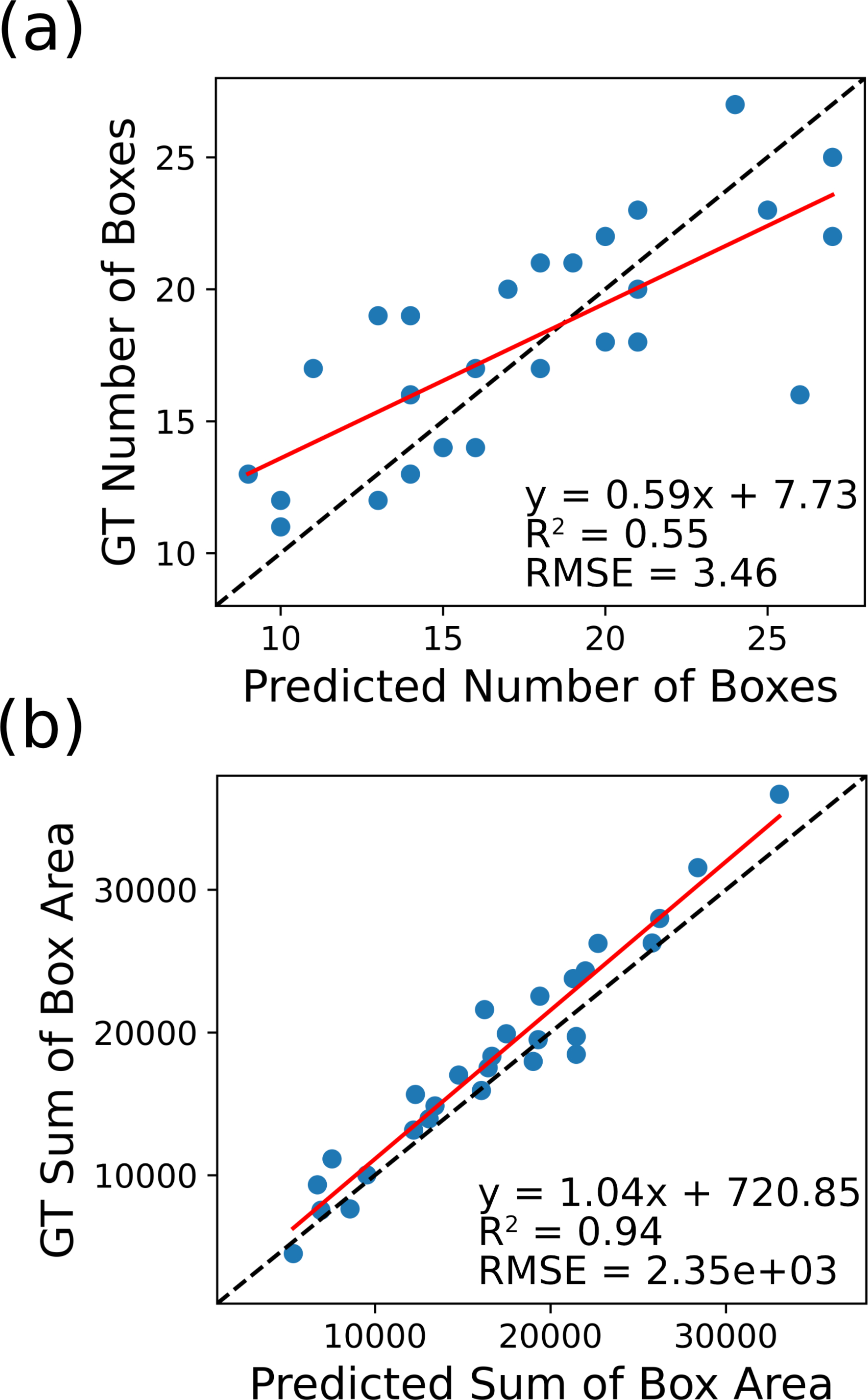}
  \caption{Internal validation of YOLOv5 model on A) box count and B) summed box area hand-labeled test set of 28 images. The red line represents the linear fit to the data and the dashed black line represents 1:1 accordance.}
  \label{fig:fig4}
\end{wrapfigure}

Within the 28 labeled test images, the model achieved an AP score of 0.56, an R2 of 0.55, and a RMSE of 3.5 bunches in prediction of the number of grape bunches in each image (Fig. \ref{fig:fig4}). Prediction of summed bounding box area demonstrated an R2 of 0.94. 
While many previous studies have focused on berry counting, some have examined bunch detection using the YOLO model. Sozzi et al. \cite{Sozzi2022} demonstrated an AP value of 0.76 with YOLOv5. However, grape bunches visualized in the study appear in higher resolution, with less occlusion than seen in the present work. Li et al. \cite{Li2021} designed a modified YOLOv4 model with integrated attention mechanisms, soft non-maximum suppression, and depthwise-separable convolutions to achieve a high prediction AP of up to 0.9. However, comparison with this work is difficult, as the intersection over union used to calculate AP in the was not stated. Finally, Santos et al. \cite{Santos2020} used CNNs for grape detection in vineyards, including YOLOv3 object detection and Mask-RCNN segmentation models. The AP score achieved for YOLOv3 at 0.5 IOU was 0.39, which is considerably lower than the YOLOv5 score in this study, although the YOLOv2 model in the previous work achieved a score of 0.48, which is similar to this study. In this previous study, Mask-RCNN achieved an AP of 0.72, representing a considerable improvement. However, Mask R-CNN is a much more computationally expensive network than YOLOv5.

\begin{figure}
  \centering
  \includegraphics[width=1.0\textwidth]{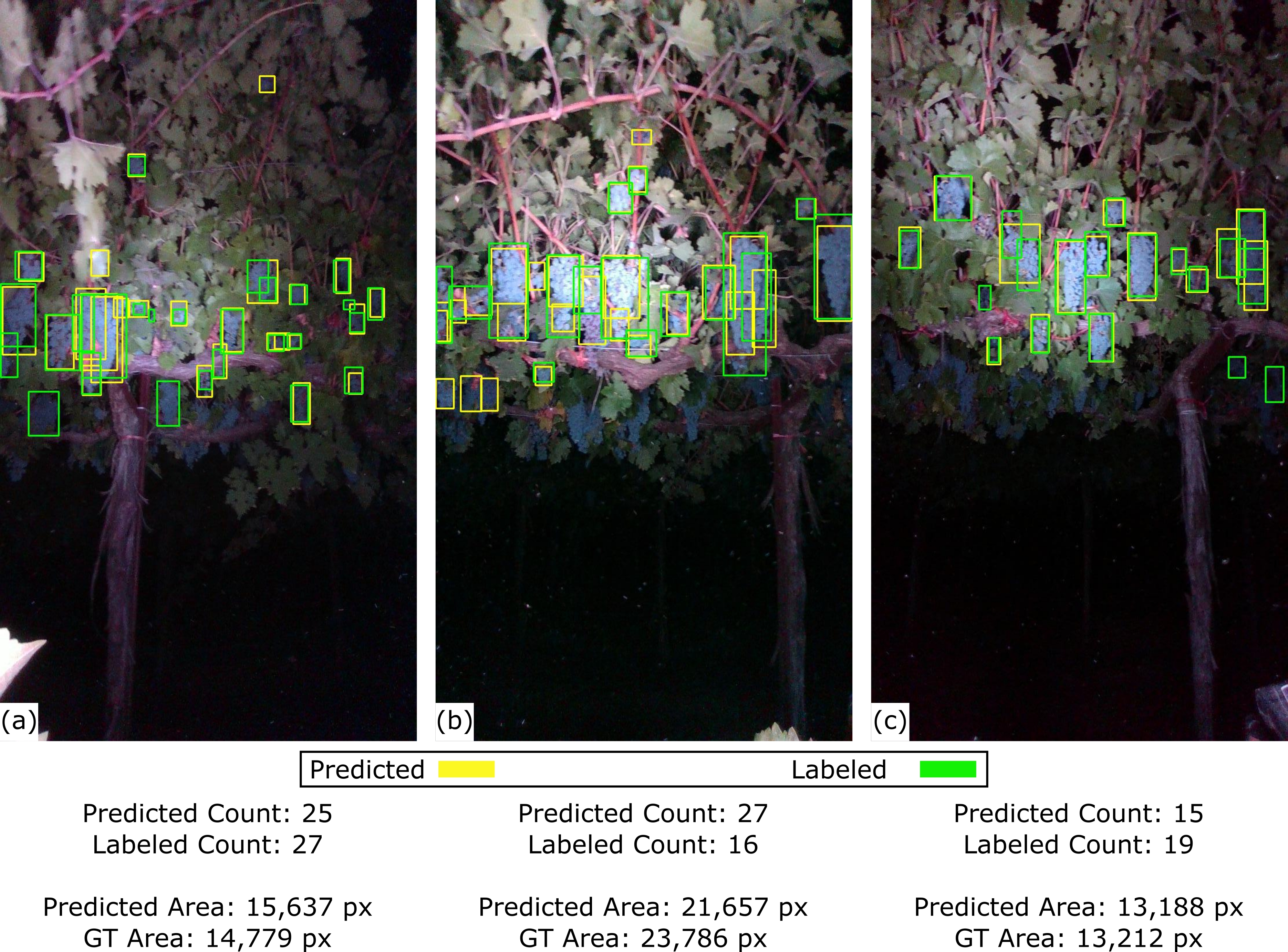}
  \caption{Example images with predicted (yellow) and labeled (green) grape bunches. Only bunches on the near side of the vine were labeled. Quantitative measures of grape yield in bunch count and area are displayed below each frame. Each frame represents the following scenario: a) missed bunches (lower left) b) extra bunches (lower left) c) missed bunches (lower right). In all instances, predicted area aligned well with ground truth area.}
  \label{fig:fig5}
\end{figure}

While the AP metric is a valuable indicator of model performance, it may not represent performance relevant to yield estimation. Notably, inconsistencies in bunch counts amounted to an RMSE of 3.46 bunches, primarily due to overcounting in the model. However, the degree of counting error was inconsistent, leading to an R2 of 0.55. This inconsistency may have been due to clustering of bunches, resulting in multiple bunches classified as one, which is a common issue \cite{DiGennaro2019,Liu2017b}. Moreover, additional grape bunches missed during labeling were detected by the model in some instances (Fig. \ref{fig:fig5}). These bunches may have constituted those on the far side of the vine (Section 2.4.1), such as in Fig. \ref{fig:fig5}b, where the predicted count is much higher than the labeled count. However, although the raw value of bunch counts was not always accurately predicted, the correlation between measured and predicted summed box area is strong, with an R2 of 0.94 (Fig \ref{fig:fig4}b). With an area-based approach, two labeled bunches counted as one as well as one labeled bunch split into multiple individual bunches may not influence the predicted area, even when the count is affected. These results suggest that in the case of box area, the object detection model is consistent with human-labeled fruit annotations. A sample of the images and labels used to train the YOLO model in this study was added as the “grape\_detection\_californianight” dataset within the open-source AgML library (\url{https://github.com/Project-AgML/AgML}).

\subsubsection{Yield Estimation using Object Detection}
\begin{figure}
  \centering
  \includegraphics[width=1.0\textwidth]{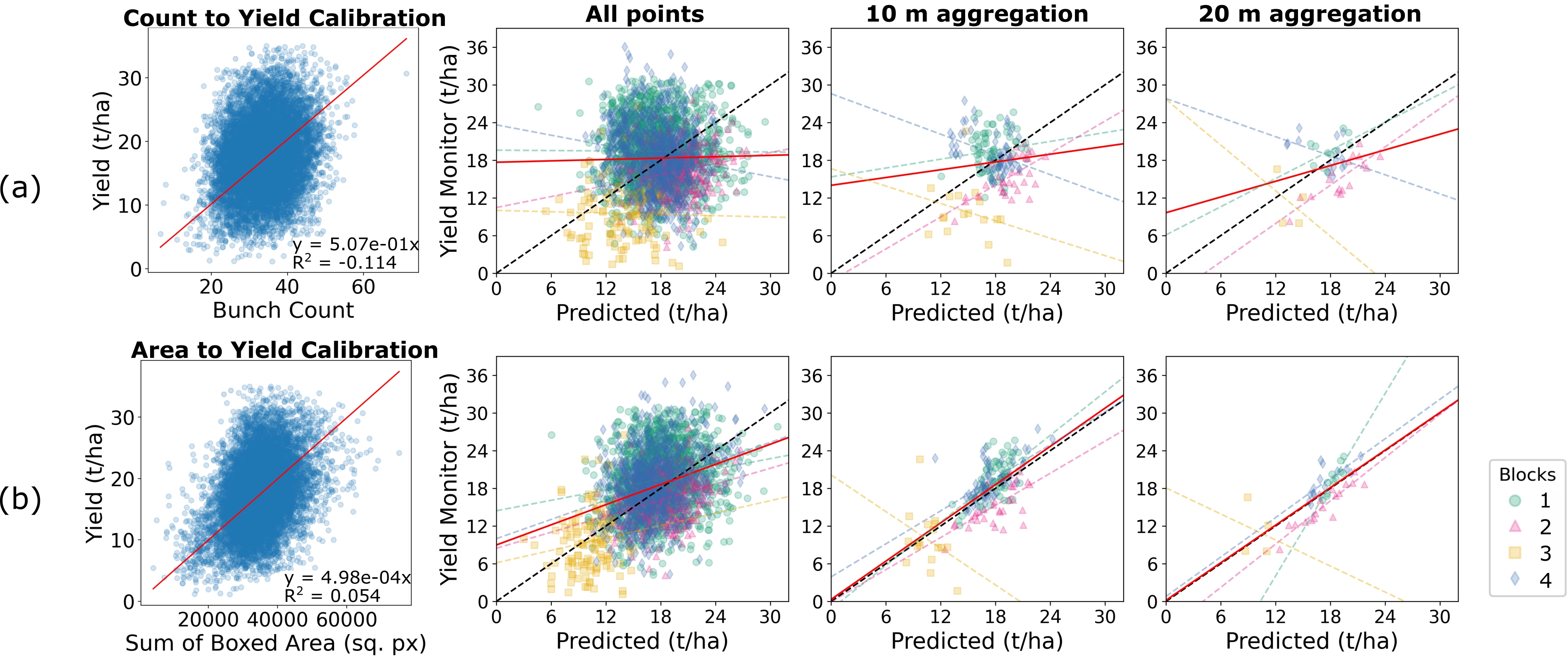}
  \caption{Yield estimation on data from the test set using the object detection model with both a) box count and b) summed box area. The first pane represents the calibration performed using the training set data to translate bunch count or box area to ground truth yield value in tons per hectare (t/ha). The red line in each pane represents the model fit to all data points in the pane. Each dashed line represents the model fit to each block of corresponding color. Black dotted lines represent the 1:1 line. The level of spatial aggregation is shown above each column after the calibration results (first column). }
  \label{fig:fig6}
\end{figure}

To predict yield with the object detection model, correlations were made between bunch count and ground truth yield as well as boxed area and ground truth yield (Fig. \ref{fig:fig6}). Using the training set, the linear relationship between count and t/ha demonstrated a poor fit, with an R2 of -0.11. Consequently, use of the bunch count model with this relationship showed poor accordance with yield in the test set with a RMSE of 6.27 t/ha and MAPE of 34\% (Table \ref{tab:table2}).

This result is similar to previous work, as studies incorporating bunch detection steps have typically used the bunch counts along with berry counts due to the variance in size between bunches \cite{DeLaFuente2015,Nuske2014} or used pixel area to express the difference between bunch sizes \cite{DiGennaro2019}. Nevertheless, block 3 appears isolated with overall lower yield predicted relative to other blocks. This may have been due to the bilateral training method, which was only used in block 3 (Table \ref{tab:table1}). Although the overall prediction performance is poor, the distinction between block 3 and the others is still apparent, suggesting that the lowered yield is related in some part to a lower bunch count.

Like in the bunch count approach, yield estimates using bounding box area require a relationship to be mapped between area and yield on the training set. This linear fit performed better than the bunch count results, but was still a poor fit overall, with an R2 of 0.054 (Fig. \ref{fig:fig6}b). This value is still lower than previously reported correlations between pixel count and grape yield, with Diago et al. \cite{Diago2012} reporting an R2 of 0.76. However, the field layout used in this previous study was comparatively simple, with a white background placed behind each vine to avoid influence from adjacent rows, increased image resolution, and a vertical shoot positioned (VSP) trellis system in which cluster occlusion was reduced. In comparison, the quadrilateral cane trellis system featured primarily in this work exhibited increased occlusion, and the images were collected without compensation for adjacent rows (see Section 3.3 below). 
Interestingly, although the relationship between summed area and yield is poor, the internal validation results demonstrated good accordance with labeled data (Fig. \ref{fig:fig4}b), demonstrating that the relationship between observed bunches and yield value is inconsistent. Detection of grape bunches on the far side of the vine may have influenced the yield estimation, although this was not the case in the labeled test set (Fig. \ref{fig:fig4}b). Therefore, the results indicated that even if a model were to accurately label all grape bunches as well as a human, the relationship to yield may still be poor. This suggests the amount of fruit present on the vine but invisible to the camera lens was not consistent throughout rows, as is sometimes assumed for the sake of modeling occlusion \cite{Bargoti2017}. 

Still, even with poor performance in relating area to yield, for predictions on the test set, performance using the box area approach was improved over the bunch count approach at all levels of spatial aggregation. Performance at the 10 m level aggregation demonstrated an R2 of 0.45 and RMSE of 3.32 t/ha. This increase in performance over the bunch count model was most likely due to accounting for variability in bunch size. Additionally, while the bunch count model was only able to express 49.7\% of the range of values at 10 m aggregation, the box area model increased this to 62.2\%. Moreover, looking at the slope and intercept of the linear fit to the test data (Table \ref{tab:table2}), the relationship is close to a 1:1 accordance, with a slope of 1.01 and intercept of 0.32 t/ha.
In addition to the high levels of occlusion in vines imaged in this study relative to previous works, one potential reason for decreased performance relative to existing object detection approaches is the small number of labeled images. Only 150 images were labeled, with 98 images used to train the model, representing 2630 total labeled bunches, and 1696 labeled bunches in the training set. For comparison, the Wine Grape Instance Segmentation Dataset (WGISD) published by Santos et al. (2020) contains 300 images with 4432 clusters. However, the dataset in the present work was distinct as the vineyards were mechanically managed, and therefore the occlusion level was much higher. Additionally, the WGISD was created as an instance segmentation dataset, as opposed to the object detection dataset used in this work. Even so, labeling images for the present study represented a considerable amount of labor as well as a bottleneck in model development. This has been noted in previous works \cite{Rahnemoonfar2017a,Santos2020}, in which the laborious nature of image labeling was specifically noted. This labeling effort becomes increasingly burdensome as the variability of images conditions increases, pointing to a need for more data efficient methods for labeling, such as in Fei et al. \cite{Fei2021}. In this work, the requirement for labeling was sidestepped in the regression approach.

\begin{table}[htbp]
  \small
  \centering
  \caption{Performance summary for each architecture on test set data only. Between the three levels of spatial aggregation, “All points” is represented by 2587 yield points, 10 m by 179 yield points, and 20 m by 47 yield points.}
    \begin{tabular}{rrrrrrrrr}
    \toprule
    \multicolumn{1}{p{2.215em}}{Architecture} & \multicolumn{1}{p{5.215em}}{Details} & \multicolumn{1}{p{6.215em}}{Spatial Aggregation (m)} & \multicolumn{1}{p{2.215em}}{RMSE (t/ha)} & \multicolumn{1}{p{2.215em}}{MAPE (\%)} & \multicolumn{1}{p{2.215em}}{R\textsuperscript{2}} & \multicolumn{1}{p{4.215em}}{Range Expressed (\%)} & \multicolumn{1}{p{3.215em}}{Fit Line Slope} & \multicolumn{1}{p{4.215em}}{Fit Line Intercept (t/ha)} \\
    \midrule
          &       & \multicolumn{1}{c}{All points} & 6.27  & 34    & 0     & 71.3  & 0.04  & 17.7 \\
          & \multicolumn{1}{c}{Bunch Count } & \multicolumn{1}{c}{10}    & 4.75  & 29.4  & 0.013 & 49.7  & 0.21  & 14 \\
    \multicolumn{1}{c}{Object Detection} &       & \multicolumn{1}{c}{20}    & 4     & 22.1  & 0.081 & 74.5  & 0.42  & 9.64 \\
\cmidrule{2-9}          &       & \multicolumn{1}{c}{All points} & 5.39  & 27.5  & 0.117 & 79.7  & 0.53  & 8.98 \\
          & \multicolumn{1}{c}{Box Area} & \multicolumn{1}{c}{10}    & 3.32  & 18.5  & 0.446 & 62.2  & 1.01  & 0.32 \\
          &       & \multicolumn{1}{c}{20}    & 2.46  & 12.2  & 0.577 & 86.3  & 1     & 0.17 \\
    \midrule
          &       & \multicolumn{1}{c}{All points} & 5.06  & 26.9  & 0.136 & 44.7  & 0.87  & 2.81 \\
    \multicolumn{1}{c}{CNN} &   \multicolumn{1}{c}{--}    & \multicolumn{1}{c}{10}    & 3.16  & 18.7  & 0.51  & 49    & 1.3   & -5.15 \\
          &       & \multicolumn{1}{c}{20}    & 2.78  & 15.1  & 0.501 & 63.2  & 1.2   & -4.13 \\
    \midrule
          &       & \multicolumn{1}{c}{All points} & 5.26  & 28    & 0.096 & 56.5  & 0.61  & 7.37 \\
          & \multicolumn{1}{c}{No Metadata} & \multicolumn{1}{c}{10}    & 3.47  & 19.6  & 0.376 & 57.7  & 0.97  & 0.54 \\
    \multicolumn{2}{l}{Transformer} & \multicolumn{1}{c}{20}    & 3.07  & 16.5  & 0.371 & 79.9  & 0.97  & -0.2 \\
          &   \multicolumn{1}{c}{Position,}    & \multicolumn{1}{c}{All points} & 5.12  & 27.1  & 0.149 & 60    & 0.64  & 6.72 \\
          & \multicolumn{1}{c}{Orientation} & \multicolumn{1}{c}{10}    & 3.24  & 18    & 0.46  & 63.8  & 0.91  & 1.73 \\
          &    \multicolumn{1}{c}{Information}   & \multicolumn{1}{c}{20}    & 2.89  & 15.2  & 0.43  & 80.5  & 0.85  & 2.42 \\
    \bottomrule
    \end{tabular}%
  \label{tab:table2}%
\end{table}%

\subsection{Regression Approaches}
\begin{figure}
  \centering
  \includegraphics[width=1.0\textwidth]{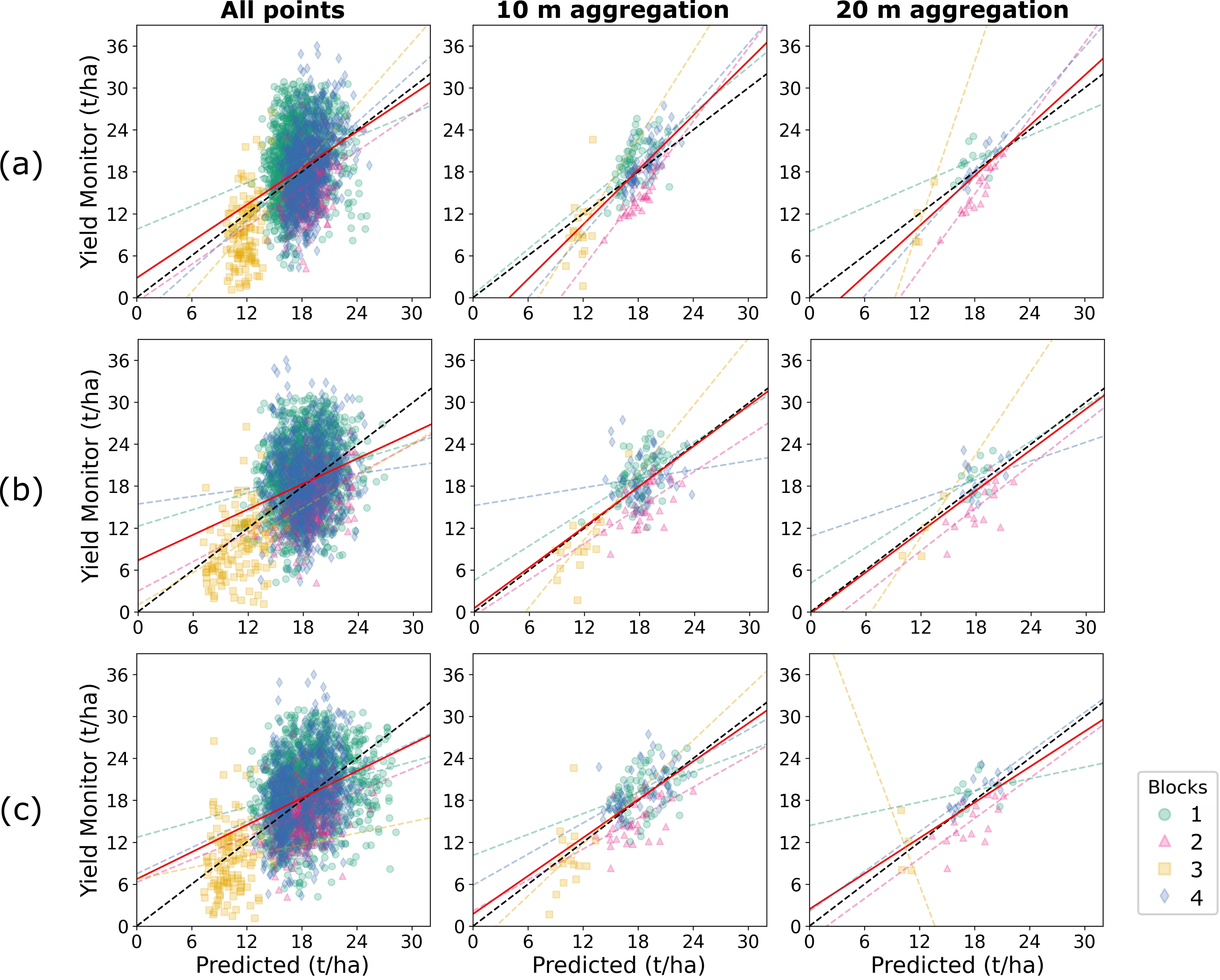}
  \caption{Yield estimation on data from the test set using the object detection model with both a) box count and b) summed box area. The first pane represents the calibration performed using the training set data to translate bunch count or box area to ground truth yield value in tons per hectare (t/ha). The red line in each pane represents the model fit to all data points in the pane. Each dashed line represents the model fit to each block of corresponding color. Black dotted lines represent the 1:1 line. The level of spatial aggregation is shown above each column after the calibration results (first column). }
  \label{fig:fig7}
\end{figure}

Unlike the object detection approach, each of the regression models were trained end-to-end on yield from images as input, which resulted in an output of yield in units of t/ha. Additionally, no manual labeling was required after alignment of yield points with image locations (Fig. \ref{fig:fig3}). Predicted yield from each of these approaches once again highlights the distinction between block 3 and the other blocks (Table \ref{tab:table1}), and block 3 yield points appear more isolated than in the object detection results. Unlike the object detection models, which were trained to detect grape bunches exclusively, the regression approaches may have used other features of the images of the bilateral trellis to distinguish between block 3 and the others. Relative to the object detection models, the CNN model further increased performance in yield estimation at 10 m spatial aggregation (Fig. \ref{fig:fig7}a), with an RMSE of 3.16 t/ha, MAPE of 18.7\%, and R2 of 0.51 (Table \ref{tab:table2}). However, the model was only able to express 49\% of the measured range at 10 m aggregation. Additionally, the line of best fit demonstrates bias, with a slope of 1.3 and intercept of -5.15 t/ha.

The transformer model without included metadata demonstrated similar performance to the CNN model (Fig. \ref{fig:fig7}b), with an R2 of 0.38, MAPE of 19.6\%, and RMSE of 3.47 t/ha at 10 m aggregation (Table \ref{tab:table2}). While the R2 and other error metrics were lower, the line of best fit represented close to 1:1 accordance, with a slope of 0.97 and an intercept of 0.54 t/ha. Additionally, range expressed by the transformer model demonstrated a slight improvement over the CNN approach, with 58\% vs. 49\% in the CNN. This flexibility to predict values at the extents of the measured range may be a result of the increased context of the input data combined with the attention mechanism of the model architecture. However, it is notable that the box area model achieved an increased level of range expressed compared to the CNN and transformer model without positional metadata, with the box area model showing 62.2\% range expressed compared with 49\% and 58\% expressed by the CNN and transformer models, respectively.
However, when positional metadata was added to the transformer input, range expressed as well as MAPE was improved over all other architectures (Fig. \ref{fig:fig7}c), with an R2 of 0.46, MAPE of 18.0\%, and RMSE of 3.24 t/ha at 10 m aggregation (Table \ref{tab:table2}). However, the line of fit was slightly further away from 1:1, with a slope of 0.91 and intercept of 1.73 t/ha. However, the range expressed by the model increased over the other end-to-end approaches, with 64\% expressed at 10 m aggregation. Since the only modification relative to the other transformer model was the addition of positional metadata, this data likely allowed the model to better attend to the input data and selectively map the inputs to higher and lower yield values. The addition of positional metadata to deep learning models in the agricultural domain has been previously demonstrated by Bargoti and Underwood \cite{Bargoti2015,Bargoti2017} in the study of image segmentation in apple orchards. Along with input image patches, these previous models were designed to accept metadata including pixel positions, row numbers, and solar position. This additional information was shown to improve performance of a multi-scale multi-layered perceptron (MLP) segmentation model, but negligibly impacted performance of a CNN model. This may have been due to the interaction between the added data and the model architectures. In the present work, the metadata included with the transformer model was added before the transformer encoder layers, as opposed to the CNN model in the previous work, which added metadata in one of the last layers \cite{Bargoti2017}.
Of all models, the best performance by MAPE (as well as range expressed) at 10 m aggregation was achieved by the transformer model with positional metadata, with an MAPE of 18\% and 63.8\% range expressed. However, at 20 m aggregation, the box area model achieved the best results, with a MAPE of 12\%. Range expressed by the box area model was also the highest of all models at 20 m aggregation, with 86\%. However, it should be noted that the box area model could only be trained after labeling 150 images of grape clusters, whereas the end-to-end models required no labeling. Moreover, the architecture of the end-to-end model is flexible towards the addition of metadata, which increases the potential for integration of proximal imagery models with other data sources, such as remote sensing data, for the improvement of predictive performance.

\subsection{Performance Relative to Previous Works}
The model performance at varying spatial aggregation values can be compared with previous remote sensing studies, such as Sun et al. \cite{Sun2017}, in which vineyard yield in a location in the California Central Valley, similar to the one in this work was predicted at 30 m using NDVI and LAI. In their work, they achieved up to 5.9 – 14.8\% error, although the authors did not split the data into training and testing sets, so the error is correlation error instead of prediction error. Additionally, that value represented the best possible correlation among many possible combinations of cumulative vegetation index maps created across the season, which the authors noted cannot be known a priori. 
At 20 m spatial aggregation (selected as the most similar aggregation level to 30 m satellite models), the regression models in this study performed well, with the transformer model with positional metadata demonstrating 2.89 t/ha RMSE and 15\% error. However, performance evaluated without spatial aggregation was worse, with an RMSE of 5.12 t/ha and 27.1\% error (Table \ref{tab:table2}). The values achieved without spatial aggregation are considerably lower than those of previous works, with Nuske et al. \cite{Nuske2014} demonstrating a relationship between detected berry count and yield with an R2 of between 0.6 and 0.73 on an individual vine level. Other authors have obtained similar results with vine-level relationships with R2 values > 0.7 \cite{Diago2012,Millan2018}. However, previous studies on yield estimation have focused on unsupervised computer vision techniques, such as use of keypoint detection \cite{Nuske2014} or distance-based metrics performed on color data \cite{Millan2018}. In these previous approaches, models were developed based on the appearance of images which were also used to generate performance metrics, as opposed to this study, which implemented a representative holdout test set. For example, one previous study on yield estimation from UAV imagery of grape canopies used images from one harvest year to estimate yield in the following year. However, according to the authors, images used for performance evaluation were selected as those with the best conditions, as opposed to a representative sample \cite{DiGennaro2019}. The previous regression CNN study by Silver and Monga \cite{Silver2019} implemented a cross-validation approach in which images were divided into five folds and five models were trained with each fold held out. However, data for that study were collected with considerable manual intervention, both in the field and via manual image processing. As such, the results are not comparable with this study and previous proximal imaging studies which allow for yield estimation on a large scale \cite{DiGennaro2019,Millan2018,Nuske2014}.
Therefore, a better comparison with correlative remote sensing as well as unsupervised proximal imaging approaches may be the performance achieved on data used to train the model (Table \ref{tab:table3}). In that context, models in the current work perform very well, with the transformer model with positional metadata achieving an R2 of 0.91 at 20 m spatial aggregation. Notably, however, results without spatial aggregation are still low, with an R2 of 0.54 in the same transformer model.

\begin{table}[htbp]
  \centering
  \caption{Performance summary for each architecture on training set data only. Between the three levels of spatial aggregation, “All points” is represented by 9509 yield points for the object detection and CNN models and 15024 yield points for the transformer models. 10 m is represented by 577 yield points for the object detection and CNN models, and by 617 for the transformer models. Finally, 20 m is represented by 171 yield points for the object detection and CNN models, and 182 points for the transformer models.}
    \begin{tabular}{rrrrrrrrr}
    \toprule
    \multicolumn{1}{p{2.215em}}{Architecture} & \multicolumn{1}{p{5.215em}}{Details} & \multicolumn{1}{p{6.215em}}{Spatial Aggregation (m)} & \multicolumn{1}{p{2.215em}}{RMSE (t/ha)} & \multicolumn{1}{p{2.215em}}{MAPE (\%)} & \multicolumn{1}{p{2.215em}}{R\textsuperscript{2}} & \multicolumn{1}{p{4.215em}}{Range Expressed (\%)} & \multicolumn{1}{p{3.215em}}{Fit Line Slope} & \multicolumn{1}{p{4.215em}}{Fit Line Intercept (t/ha)} \\
    \midrule
          &       & \multicolumn{1}{c}{All points} & 6.12  & 34.2  & 0.044 & 96.7  & 0.35  & 11.65 \\
          & Bunch Count  & \multicolumn{1}{c}{10}    & 4.05  & 20.3  & 0.242 & 58.3  & 0.79  & 4.17 \\
    \multicolumn{1}{l}{Object Detection} &       & \multicolumn{1}{c}{20}    & 3.62  & 16.8  & 0.254 & 57.2  & 0.66  & 6.62 \\
\cmidrule{2-9}          &       & \multicolumn{1}{c}{All points} & 5.64  & 30.7  & 0.129 & 104   & 0.57  & 7.75 \\
          & Box Area & \multicolumn{1}{c}{10}    & 3.36  & 15.5  & 0.477 & 74.6  & 0.95  & 1.41 \\
          &       & \multicolumn{1}{c}{20}    & 2.74  & 12.4  & 0.52  & 79.9  & 0.91  & 1.91 \\
    \midrule
          &       & \multicolumn{1}{c}{All points} & 4.73  & 27.2  & 0.342 & 55.2  & 1.15  & -2.59 \\
    \multicolumn{1}{l}{CNN} & --    & \multicolumn{1}{c}{10}    & 2.49  & 12.5  & 0.732 & 55    & 1.26  & -4.63 \\
          &       & \multicolumn{1}{c}{20}    & 2.01  & 9.2   & 0.761 & 59    & 1.22  & -3.99 \\
    \midrule
          &       & \multicolumn{1}{c}{All points} & 3.93  & 23.1  & 0.568 & 84.4  & 1.04  & -0.71 \\
          & No Metadata & \multicolumn{1}{c}{10}    & 1.4   & 6.3   & 0.908 & 87.3  & 1.06  & -0.86 \\
    \multicolumn{1}{l}{Transformer} &       & \multicolumn{1}{c}{20}    & 1.13  & 4.8   & 0.932 & 86.2  & 1.07  & -0.9 \\
\cmidrule{2-9}          & Position, & \multicolumn{1}{c}{All points} & 4.09  & 23.4  & 0.535 & 95.3  & 0.93  & 1.15 \\
          & Orientation & \multicolumn{1}{c}{10}    & 1.49  & 6.4   & 0.893 & 87.8  & 0.97  & 0.67 \\
          & Information & \multicolumn{1}{c}{20}    & 1.23  & 4.9   & 0.91  & 90.7  & 0.99  & 0.49 \\
    \bottomrule
    \end{tabular}%
  \label{tab:table3}%
\end{table}%

However, in addition to the lack of a holdout set, previous studies were conducted in very different environmental conditions as compared with this work. Almost all previous studies on grape yield estimation have been performed with VSP trellis configurations and manual defoliation. As an example of the effect that trellis configuration has on proximal imaging, one previous study of image-based methods for grape phenotyping noted that in point cloud classification of grape pixels, precision and recall dropped by 39\% and 6\%, respectively, between vines trained with a VSP trellis and semi-manual pruned hedge (SMPH) trellis \cite{Rose2016}. The SMPH trellis configuration seen in that work is more similar to the quadrilateral trellis featured primarily in the present study as opposed to the VSP trellis. Additionally, Di Gennaro et al. \cite{DiGennaro2019} separated vine images into conditions based on vigor as well as occlusion and found the true positive rate of pixel classification was reduced by 45 and 73\% in high and low vigor vines, respectively, between good conditions representing minimal occlusion and poor conditions representing occlusion and shading. On the contrary, this work was performed in a commercial vineyard within the California Central Valley, where due to the heat and dry conditions, fruit shading is extremely important, and VSP trellis configurations and defoliation would be detrimental to grape quality. Moreover, vines in this study were mechanically trimmed and harvested, which is becoming more common in the wine grape industry \cite{KaanKurtural2021}. As a result, robust models which work reliably in vineyards managed mechanically will become more important in the future. Additionally, previous work has relied on hand-weighed yield data collected at the vine level, which is typically not implemented by commercial growers on a large scale. Instead, yield monitor data is more commonly used \cite{Bramley2004}, which allows for a larger dataset but also introduces additional sources of error into the yield estimates due to both the continuous nature of data collection as well as the influence of yield monitor geometry and mass flow on the measurement \cite{Searcy1989}. Additionally, the use of larger datasets requires more consideration placed on the efficiency of data collection and processing. In this study, collection of 274,944 images from approximately 13,180 vines required approximately 5 hours for data collection (images were recorded at 15 Hz), and approximately 32 GB of data in JPEG format, which needed to be transferred from the sensing kit to a server for further use. This data was used for training the CNN and transformer models, which typically benefit from a large quantity of data \cite{Dhillon2020}. However, after models are trained, deployment of sensing kits with edge computing capability may allow for real-time model prediction, where image data is discarded after predictions are made, reducing the storage and bandwidth requirements for data handling. This strategy was outlined recently by Guillén et al \cite{Guillen2021}, where an NVIDIA Jetson single board computer, similar to the device present in this study, was used to make offline predictions of temperature in an agricultural environment with limited bandwidth. Furthermore, mounting of sensing kits on equipment during field operations typically performed at night, such as sulfur application, would eliminate the need for time dedicated to imaging along, greatly improving efficiency of data collection or vineyard monitoring. 

\subsection{Visualization and Saliency Mapping}
\begin{figure}
  \centering
  \includegraphics[width=1.0\textwidth]{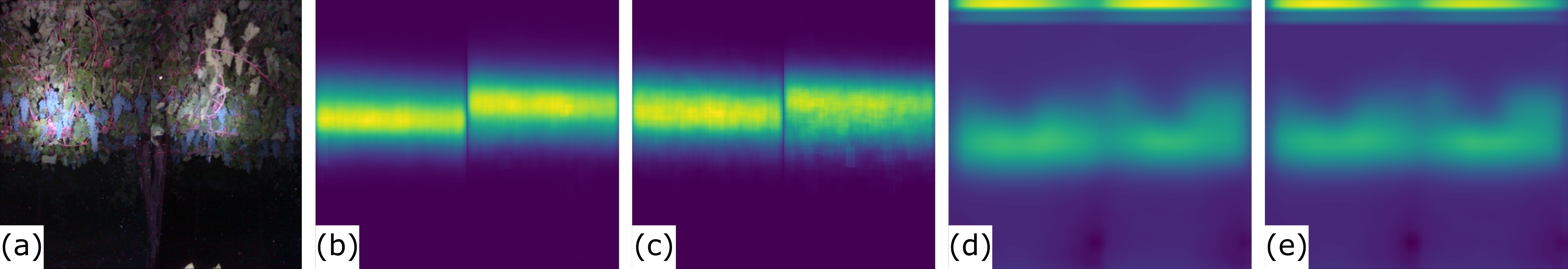}
  \caption{(a) Example input image to the CNN model. (b) Heatmap representing localization of predicted grape bunches using the object detection model across the train set and c) test set. (d) Grad-CAM heatmap representing areas averaged over all test set yield points with high influence on increasing predicted yield in the train set and e) test set.}
  \label{fig:fig8}
\end{figure}

Figure \ref{fig:fig8} contains an example input image to the CNN model, along with heatmaps representing the position of detected grapes on the vine in the training (Fig. \ref{fig:fig8}b) test set (Fig \ref{fig:fig8}c) as well as the average Grad-CAM response from the CNN network for the training and test sets (Fig \ref{fig:fig8}d-e). 

In this visualization, although grape clusters are localized in the center of the image frame, the regions of the image with the strong contribution to predicted yield include the center of the frame as well as the top of the frame, potentially representing the density of the vine canopy at the top of the frame, where in some images, dark background is visible through the canopy. The fact that the CNN model was free to use these features of the input, whereas the object detection model was constrained to regions with grape bunches, may have contributed to the increased performance of both the CNN and transformer models over the object detection models. Alternatively, the attention on the top edge may also be a result of overfitting to the dataset by relating spurious image features with yield. This behavior was seen in both the training and test sets (Fig. \ref{fig:fig8}d-e), suggesting that the attention was learned by the model, and not due to differences between the datasets.

\begin{figure}
  \centering
  \includegraphics[width=1.0\textwidth]{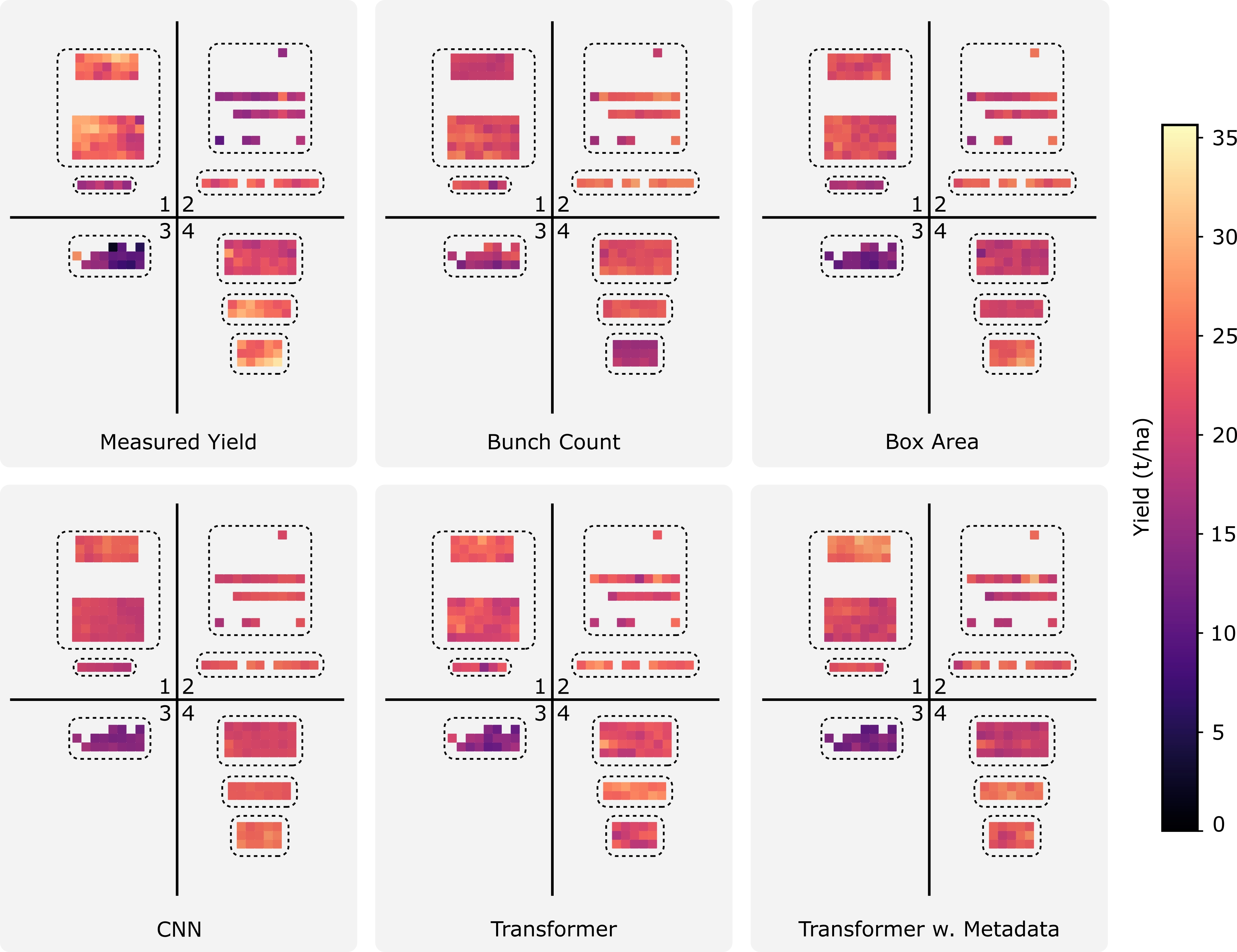}
  \caption{Yield maps from test set yield points derived from all data sources (represented by dark points in Fig. \ref{fig:fig1}). Measured yield represents the ground truth yield from the yield monitor. Each additional map represented predicted test set yield using a different model. Maps were generated by aggregating data to 10m. Numbers in each pane represent block numbers, as in Fig. \ref{fig:fig1}.}
  \label{fig:fig9}
\end{figure}

Yield maps generated from measured and predicted data demonstrate how the models tend to predict values close to the center of the distribution of values (Fig. \ref{fig:fig9}). This is a visual representation of the range expressed results in Table \ref{tab:table2}, which demonstrate that no predicted dataset was able to achieve the same range of values as the measured data (< 100\% range expressed). Nevertheless, visual trends from the measured plots can be seen in the predicted maps. Most notably, block 3 is represented on all maps with the lowest yield values. Likewise, the pattern shown in block 2 by both transformer models is similar to that of the measured data. Future work will involve collecting more data from more contiguous regions such that more detailed maps may be generated with the proximal imaging approach used in this work.

\section{Conclusions}
Three different models for prediction of grape yield from proximal imagery were trained based on image data collected from a vehicle-mounted sensing kit against ground truth yield data collected from a mechanical yield harvester. The object detection model, which is most similar to previous methods of yield estimation from proximal imagery, demonstrated poor performance when the size of grape bunches was not accounted for (MAPE of 29.4\% when aggregated to 10 m blocks). Performance improved with use of grape bunch area as opposed to bunch count, with a MAPE of 18.5\% at 10 m spatial aggregation. 
Regression-based deep learning models trained end-to-end on yield prediction from input imagery demonstrated similar performance to the best object detection approach (grape bunch area). Regression CNN and transformer architectures were utilized, with an MAPE of up to 18\% at 10 m spatial aggregation achieved with a transformer architecture after the addition of encoded positional metadata. Moreover, these regression architectures eliminate the need for hand-labeling images, removing a considerable bottleneck from the model development process.
While the performance on the test set in this work is lower than some previous studies, previous works have primarily been performed in vineyards trimmed by hand with vertical shoot positioned trellis configurations. This study represents an application of yield prediction in challenging conditions for image collection, with high occlusion, dense foliage, and a quadrilateral trellis configuration, common to the California Central Valley. Additionally, this study assessed performance on a holdout test set, which is more representative of unseen data.
Future work in this area will encompass yield forecasting from earlier in the season, at times prior to veraison, where accurate yield estimates are difficult to obtain but highly valued. Additionally, while the image data in this work were collected specifically for this study, the low-cost, vehicle-mounted sensing kit used for imaging may allow for automated data collection during routine management operations in the future, if the kit is mounted on existing equipment. Finally, due to the flexibility of the regression-based approach, data fusion techniques where proximal imagery is supplemented with remote sensing data represents a promising area for exploration in which coarse resolution satellite data may be used to improve performance of models trained on ground-based imagery.

\section{CRediT Authorship Contribution Statement}
\textbf{Alexander G. Olenskyj}: Conceptualization, Methodology, Validation, Formal Analysis, Data Curation, Writing – Original Draft, Writing – Review \& Editing, Visualization \textbf{Brent S. Sams}: Conceptualization, Resources, Data Curation, Writing – Review \& Editing \textbf{Zhenghao Fei}: Resources, Software, Data Curation, Writing – Review \& Editing \textbf{Vishal Singh}: Resources, Project Administration \textbf{Pranav V. Raja}: Data Curation \textbf{Gail M. Bornhorst}: Conceptualization, Writing – Review \& Editing, Supervision \textbf{J. Mason Earles}: Conceptualization, Investigation, Resources, Writing – Review \& Editing, Supervision, Project Administration, Funding Acquisition

\section{Acknowledgements}
This project was partly supported by the USDA AI Institute for Next Generation Food Systems (AIFS), award \#2020-67021-32855, and by the NSF-funded UC Davis Center for Data Science and Artificial Intelligence, award \#1934568. AO was partly funded by the American Society for Enology and Viticulture Michael Vail Scholarship. The authors would also like to acknowledge E. \& J. Gallo Winery for their support in enabling field data collection in their vineyards as well as for providing the yield monitor data used in this project.

\bibliographystyle{unsrt}  
\bibliography{template}  

\begin{thebibliography}{10}

\bibitem{Diago2012}
Maria~Paz Diago, Christian Correa, Borja Mill{\'{a}}n, Pilar Barreiro,
  Constantino Valero, and Javier Tardaguila.
\newblock {Grapevine yield and leaf area estimation using supervised
  classification methodology on RGB images taken under field conditions}.
\newblock {\em Sensors (Switzerland)}, 12(12):16988--17006, 2012.

\bibitem{Nuske2014}
Stephen Nuske, Kyle Wilshusen, Supreeth Achar, Luke Yoder, Srinivasa
  Narasimhan, and Sanjiv Singh.
\newblock {Automated Visual Yield Estimation in Vineyards}.
\newblock {\em J. Field Robotics}, 31(5):837--860, 2014.

\bibitem{Liu2020}
Scarlett Liu, Xiangdong Zeng, and Mark Whitty.
\newblock {A vision-based robust grape berry counting algorithm for fast
  calibration-free bunch weight estimation in the field}.
\newblock {\em Computers and Electronics in Agriculture}, 173(December
  2019):105360, 2020.

\bibitem{Khaliq2019}
Aleem Khaliq, Lorenzo Comba, Alessandro Biglia, Davide {Ricauda Aimonino},
  Marcello Chiaberge, and Paolo Gay.
\newblock {Comparison of satellite and UAV-based multispectral imagery for
  vineyard variability assessment}.
\newblock {\em Remote Sensing}, 11(4), 2019.

\bibitem{Yang2019}
Qi~Yang, Liangsheng Shi, Jinye Han, Yuanyuan Zha, and Penghui Zhu.
\newblock {Deep convolutional neural networks for rice grain yield estimation
  at the ripening stage using UAV-based remotely sensed images}.
\newblock {\em Field Crops Research}, 235(February):142--153, 2019.

\bibitem{DiGennaro2019}
Salvatore~Filippo {Di Gennaro}, Piero Toscano, Paolo Cinat, Andrea Berton, and
  Alessandro Matese.
\newblock {A low-cost and unsupervised image recognition methodology for yield
  estimation in a vineyard}.
\newblock {\em Frontiers in Plant Science}, 10(May):1--13, 2019.

\bibitem{Gongal2015}
A.~Gongal, S.~Amatya, M.~Karkee, Q.~Zhang, and K.~Lewis.
\newblock {Sensors and systems for fruit detection and localization: A review}.
\newblock {\em Computers and Electronics in Agriculture}, 116:8--19, 2015.

\bibitem{Dunn2004}
Gregory~M. Dunn and Stephen~R. Martin.
\newblock {Yield prediction from digital image analysis: A technique with
  potential for vineyard assessments prior to harvest}.
\newblock {\em Australian Journal of Grape and Wine Research}, 10(3):196--198,
  2004.

\bibitem{Mu2020}
Yue Mu, Tai~Shen Chen, Seishi Ninomiya, and Wei Guo.
\newblock {Intact detection of highly occluded immature tomatoes on plants
  using deep learning techniques}.
\newblock {\em Sensors (Switzerland)}, 20(10):1--16, 2020.

\bibitem{Wang2021}
Yiding Wang, Yuxin Qin, and Jiali Cui.
\newblock {Occlusion Robust Wheat Ear Counting Algorithm Based on Deep
  Learning}.
\newblock {\em Frontiers in Plant Science}, 12(June):1--14, 2021.

\bibitem{Bargoti2017}
Suchet Bargoti and James~P. Underwood.
\newblock {Image Segmentation for Fruit Detection and Yield Estimation in Apple
  Orchards}.
\newblock {\em Journal of Field Robotics}, 34(6):1039--1060, 2017.

\bibitem{Gene-Mola2019}
Jordi Gen{\'{e}}-Mola, Ver{\'{o}}nica Vilaplana, Joan~R. Rosell-Polo,
  Josep-Ramon Morros, Javier Ruiz-Hidalgo, and Eduard Gregorio.
\newblock {Multi-modal deep learning for Fuji apple detection using RGB-D
  cameras and their radiometric capabilities}.
\newblock {\em Computers and Electronics in Agriculture}, 162:689--698, 2019.

\bibitem{Santos2020}
Thiago~T. Santos, Leonardo~L. de~Souza, Andreza~A. dos Santos, and Sandra
  Avila.
\newblock {Grape detection, segmentation, and tracking using deep neural
  networks and three-dimensional association}.
\newblock {\em Computers and Electronics in Agriculture}, 170:1--22, 2020.

\bibitem{Hani2020}
Nicolai H{\"a}ni, Pravakar Roy, and Volkan Isler.
\newblock A comparative study of fruit detection and counting methods for yield
  mapping in apple orchards.
\newblock {\em Journal of Field Robotics}, 37(2):263--282, 2020.

\bibitem{Maldonado2016}
Walter Maldonado and Jos{\'{e}}~Carlos Barbosa.
\newblock {Automatic green fruit counting in orange trees using digital
  images}.
\newblock {\em Computers and Electronics in Agriculture}, 127:572--581, 2016.

\bibitem{Payne2014}
A.~Payne, K.~Walsh, P.~Subedi, and D.~Jarvis.
\newblock {Estimating mango crop yield using image analysis using fruit at
  'stone hardening' stage and night time imaging}.
\newblock {\em Computers and Electronics in Agriculture}, 100:160--167, 2014.

\bibitem{Rahnemoonfar2017a}
Maryam Rahnemoonfar and Clay Sheppard.
\newblock {Deep count: Fruit counting based on deep simulated learning}.
\newblock {\em Sensors (Switzerland)}, 17(4):1--12, 2017.

\bibitem{Li2021}
Huipeng Li, Changyong Li, Guibin Li, and Lixin Chen.
\newblock {A real-time table grape detection method based on improved
  YOLOv4-tiny network in complex background}.
\newblock {\em Biosystems Engineering}, 212:347--359, 2021.

\bibitem{Liu2017b}
Scarlett Liu, Steve Cossell, Julie Tang, Gregory Dunn, and Mark Whitty.
\newblock {A computer vision system for early stage grape yield estimation
  based on shoot detection}.
\newblock {\em Computers and Electronics in Agriculture}, 137:88--101, 2017.

\bibitem{Liu2015b}
Scarlett Liu and Mark Whitty.
\newblock {Automatic grape bunch detection in vineyards with an SVM
  classifier}.
\newblock {\em Journal of Applied Logic}, 13(4):643--653, 2015.

\bibitem{Sozzi2022}
Marco Sozzi, Silvia Cantalamessa, Alessia Cogato, Ahmed Kayad, and Francesco
  Marinello.
\newblock {Automatic Bunch Detection in White Grape Varieties Using YOLOv3,
  YOLOv4, and YOLOv5 Deep Learning Algorithms}.
\newblock {\em Agronomy}, 12(319), 2022.

\bibitem{Millan2018}
Borja Millan, Santiago Velasco-Forero, Arturo Aquino, and Javier Tardaguila.
\newblock {On-the-go grapevine yield estimation using image analysis and
  boolean model}.
\newblock {\em Journal of Sensors}, 2018, 2018.

\bibitem{Rose2016}
Johann~Christian Rose, Anna Kicherer, Markus Wieland, Lasse Klingbeil, Reinhard
  T{\"{o}}pfer, and Heiner Kuhlmann.
\newblock {Towards automated large-scale 3D phenotyping of vineyards under
  field conditions}.
\newblock {\em Sensors (Switzerland)}, 16(12):1--25, 2016.

\bibitem{Milella2019}
Annalisa Milella, Roberto Marani, Antonio Petitti, and Giulio Reina.
\newblock {In-field high throughput grapevine phenotyping with a consumer-grade
  depth camera}.
\newblock {\em Computers and Electronics in Agriculture}, 156(November
  2018):293--306, 2019.

\bibitem{Hu2019}
Junjie Hu, Yan Zhang, and Takayuki Okatani.
\newblock {Visualization of convolutional neural networks for monocular depth
  estimation}.
\newblock {\em arXiv}, pages 3869--3878, 2019.

\bibitem{Ege2017}
Takumi Ege and Keiji Yanai.
\newblock {Simultaneous estimation of food categories and calories with
  multi-task CNN}.
\newblock {\em Proceedings of the 15th IAPR International Conference on Machine
  Vision Applications, MVA 2017}, pages 198--201, 2017.

\bibitem{Othmani2020}
Alice Othmani, Abdul~Rahman Taleb, Hazem Abdelkawy, and Abdenour Hadid.
\newblock {Age estimation from faces using deep learning: A comparative
  analysis}.
\newblock {\em Computer Vision and Image Understanding}, 196(April):102961,
  2020.

\bibitem{ZakirHossain2019}
M.~D. {Zakir Hossain}, Ferdous Sohel, Mohd~Fairuz Shiratuddin, and Hamid Laga.
\newblock {A comprehensive survey of deep learning for image captioning}.
\newblock {\em ACM Computing Surveys}, 51(6), 2019.

\bibitem{Silver2019}
Daniel~L. Silver and Tanya Monga.
\newblock {In Vino Veritas: Estimating Vineyard Grape Yield from Images Using
  Deep Learning}.
\newblock In {\em Canadian AI}, pages 212--224. Springer International
  Publishing, 2019.

\bibitem{Carion2020}
Nicolas Carion, Francisco Massa, Gabriel Synnaeve, Nicolas Usunier, Alexander
  Kirillov, and Sergey Zagoruyko.
\newblock {End-to-End Object Detection with Transformers}.
\newblock {\em Lecture Notes in Computer Science (including subseries Lecture
  Notes in Artificial Intelligence and Lecture Notes in Bioinformatics)}, 12346
  LNCS:213--229, 2020.

\bibitem{Dosovitskiy2020}
Alexey Dosovitskiy, Lucas Beyer, Alexander Kolesnikov, Dirk Weissenborn,
  Xiaohua Zhai, Thomas Unterthiner, Mostafa Dehghani, Matthias Minderer, Georg
  Heigold, Sylvain Gelly, Jakob Uszkoreit, and Neil Houlsby.
\newblock An image is worth 16x16 words: Transformers for image recognition at
  scale.
\newblock {\em CoRR}, abs/2010.11929, 2020.

\bibitem{Vaswani2017}
Ashish Vaswani, Noam Shazeer, Niki Parmar, Jakob Uszkoreit, Llion Jones,
  Aidan~N. Gomez, {\L}ukasz Kaiser, and Illia Polosukhin.
\newblock {Attention is all you need}.
\newblock {\em Advances in Neural Information Processing Systems},
  2017-Decem(Nips):5999--6009, 2017.

\bibitem{Xie2021}
Enze Xie, Wenhai Wang, Zhiding Yu, Anima Anandkumar, Jose~M. Alvarez, and Ping
  Luo.
\newblock Segformer: Simple and efficient design for semantic segmentation with
  transformers.
\newblock In M.~Ranzato, A.~Beygelzimer, Y.~Dauphin, P.S. Liang, and J.~Wortman
  Vaughan, editors, {\em Advances in Neural Information Processing Systems},
  volume~34, pages 12077--12090. Curran Associates, Inc., 2021.

\bibitem{Yang2004}
Chenghai Yang, James~H. Everitt, Joe~M. Bradford, and Dale Murden.
\newblock {Airborne hyperspectral imagery and yield monitor data for mapping
  cotton yield variability}.
\newblock {\em Precision Agriculture}, 5(5):445--461, 2004.

\bibitem{Sun2017}
Liang Sun, Feng Gao, Martha~C. Anderson, William~P. Kustas, Maria~M. Alsina,
  Luis Sanchez, Brent Sams, Lynn McKee, Wayne Dulaney, William~A. White,
  Joseph~G. Alfieri, John~H. Prueger, Forrest Melton, and Kirk Post.
\newblock {Daily mapping of 30 m LAI and NDVI for grape yield prediction in
  California vineyards}.
\newblock {\em Remote Sensing}, 9(4):1--18, 2017.

\bibitem{Tukey1977}
John~W. Tukey.
\newblock {\em {Exploratory Data Analysis}}.
\newblock Addison-Wesley Pub. Co., Reading, MA, 2 edition, 1977.

\bibitem{Sandler2018}
Mark Sandler, Andrew Howard, Menglong Zhu, Andrey Zhmoginov, and Liang~Chieh
  Chen.
\newblock {MobileNetV2: Inverted Residuals and Linear Bottlenecks}.
\newblock {\em Proceedings of the IEEE Computer Society Conference on Computer
  Vision and Pattern Recognition}, pages 4510--4520, 2018.

\bibitem{Jocher2021}
Glenn Jocher, Alex Stoken, Jirka Borovec, NanoCode012, ChristopherSTAN, Liu
  Changyu, Laughing, tkianai, yxNONG, Adam Hogan, lorenzomammana, AlexWang1900,
  Ayush Chaurasia, Laurentiu Diaconu, Marc, wanghaoyang0106, ml5ah, Doug,
  Durgesh, Francisco Ingham, Frederik, Guilhen, Adrien Colmagro, Hu~Ye,
  Jacobsolawetz, Jake Poznanski, Jiacong Fang, Junghoon Kim, Khiem Doan, and
  Lijun Yu.
\newblock {ultralytics/yolov5: v4.0 - nn.SiLU() activations, Weights \& Biases
  logging, PyTorch Hub integration}, January 2021.

\bibitem{He2016}
Kaiming He, Xiangyu Zhang, Shaoqing Ren, and Jian Sun.
\newblock {Deep residual learning for image recognition}.
\newblock {\em Proceedings of the IEEE Computer Society Conference on Computer
  Vision and Pattern Recognition}, 2016-Decem:770--778, 2016.

\bibitem{Nair2010}
Vinod Nair and Geoffrey~E. Hinton.
\newblock {Rectified Linear Units Improve Restricted Boltzmann Machines}.
\newblock {\em ICML}, 2010.

\bibitem{Srivastava2014}
Nitish Srivastava, Geoffrey Hinton, Alex Krizhevsky, Ilya Sutskever, and Ruslan
  Salakhutdinov.
\newblock {Dropout: A Simple Way to Prevent Neural Networks from Overfitting}.
\newblock {\em Journal of Machine Learning Research}, 15:1929--1958, 2014.

\bibitem{Barron2019}
Jonathan~T. Barron.
\newblock {A general and adaptive robust loss function}.
\newblock {\em Proceedings of the IEEE Computer Society Conference on Computer
  Vision and Pattern Recognition}, 2019-June:4326--4334, 2019.

\bibitem{Searcy1989}
S.~W. Searcy, J.~K. Schueller, Y.~H. Bae, S.~C. Borgelt, and B.~A. Stout.
\newblock {Mapping of spatially variable yield during grain combining}.
\newblock {\em Transactions of the American Society of Agricultural Engineers},
  32(3):826--829, 1989.

\bibitem{Selvaraju2016}
Ramprasaath~R. Selvaraju, Michael Cogswell, Abhishek Das, Ramakrishna Vedantam,
  Devi Parikh, and Dhruv Batra.
\newblock {Grad-CAM: Visual Explanations from Deep Networks via Gradient-based
  Localization}.
\newblock {\em ICCV}, 2016.

\bibitem{DeLaFuente2015}
Mario {De La Fuente}, Rub{\'{e}}n Linares, Pilar Baeza, Carlos Miranda, and
  Jos{\'{e}}~Ram{\'{o}}n Lissarrague.
\newblock {Comparison of different methods of grapevine yield prediction in the
  time window between fruitset and veraison}.
\newblock {\em Journal International des Sciences de la Vigne et du Vin},
  49(1):27--35, 2015.

\bibitem{Fei2021}
Zhenghao Fei, Alex Olenskyj, Brian~N. Bailey, and Mason Earles.
\newblock Enlisting 3d crop models and gans for more data efficient and
  generalizable fruit detection.
\newblock In {\em 2021 IEEE/CVF International Conference on Computer Vision
  Workshops (ICCVW)}, pages 1269--1277, 2021.

\bibitem{Bargoti2015}
Suchet Bargoti and James Underwood.
\newblock {Utilising Metadata to Aid Image Classification in Orchards}.
\newblock {\em IEEE International Conference on Intelligent Robots and Systems
  (IROS), Workshop on Alternative Sensing for Robot Perception (WASRoP)}, pages
  1--3, 2015.

\bibitem{KaanKurtural2021}
S.~{Kaan Kurtural} and Matthew~W. Fidelibus.
\newblock {Mechanization of Pruning, Canopy Management, and Harvest in
  Winegrape Vineyards}.
\newblock {\em Catalyst: Discovery into Practice}, 5(1):29--44, 2021.

\bibitem{Bramley2004}
R.~G.V. Bramley and R.~P. Hamilton.
\newblock {Understanding variability in winegrape production systems 1. Within
  vineyard variation in yield over several vintages}.
\newblock {\em Australian Journal of Grape and Wine Research}, 10(1):32--45,
  2004.

\bibitem{Dhillon2020}
Anamika Dhillon and Gyanendra~K. Verma.
\newblock {Convolutional neural network: a review of models, methodologies and
  applications to object detection}.
\newblock {\em Progress in Artificial Intelligence}, 9(2):85--112, 2020.

\bibitem{Guillen2021}
Miguel~A. Guill{\'{e}}n, Antonio Llanes, Baldomero Imbern{\'{o}}n, Raquel
  Mart{\'{i}}nez-Espa{\~{n}}a, Andr{\'{e}}s Bueno-Crespo, Juan~Carlos Cano, and
  Jos{\'{e}}~M. Cecilia.
\newblock {Performance evaluation of edge-computing platforms for the
  prediction of low temperatures in agriculture using deep learning}.
\newblock {\em Journal of Supercomputing}, 77(1):818--840, 2021.

\end{thebibliography}

\end{document}